\documentclass[journal]{IEEEtran}

\usepackage{graphicx} 
\usepackage{subfigure}
\usepackage{color}
\usepackage{multirow,subfigure,pifont,hyperref}
\usepackage{verbatim}
\usepackage{amsthm,amsmath,amssymb}
\usepackage{mathrsfs}
\usepackage[ruled]{algorithm2e}
\usepackage{setspace}
\usepackage{textcomp}
\RequirePackage{amsmath}[2016/11/05]

\begin{document}
	\title{Towards Complete-View and High-Level Pose-based Gait Recognition}
	\author{Honghu Pan, Yongyong Chen, Tingyang Xu, Yunqi He, Zhenyu He*,~\IEEEmembership{Senior Member,~IEEE}
	\thanks{This research is supported part by the National Natural Science
			Foundation of China (Grant No.62172126 and Grant No.62106063), by the Shenzhen Research Council (Grant No. JCYJ20210324120202006), by the Shenzhen College Stability Support Plan (Grant GXWD20201230155427003-20200824113231001).}
	\thanks{H. Pan, Y. Chen and Z. He are with School of Computer Science and Technology, Harbin Institute of Technology, Shenzhen, Shenzhen 518000, China. (Emails: 19B951002@stu.hit.edu.cn, YongyongChen.cn@gmail.com and zhenyuhe@hit.edu.cn)}
	\thanks{T. Xu is with Tencent AI LAB, Shenzhen 150000, China. (Email: Tingyangxu@tencent.com)}
	\thanks{Y. He is with College of Information and Computer Engineering, Northeast Forestry University, Harbin 518000, China. (Email: heyunqi.cs@gmail.com)}
	}
	
	\maketitle
	
	\begin{abstract}
		The model-based gait recognition methods usually adopt the pedestrian walking postures to identify human beings.
		However, existing methods did not explicitly resolve the large intra-class variance of human pose due to camera views changing.
		In this paper, we propose to generate multi-view pose sequences for each single-view pose sample by learning full-rank transformation matrices via lower-upper generative adversarial network (LUGAN).
		By the prior of camera imaging, we derive that the spatial coordinates between cross-view poses satisfy a linear transformation of a full-rank matrix, thereby, this paper employs the adversarial training to learn transformation matrices from the source pose and target views to obtain the target pose sequences.
		To this end, we implement a generator composed of graph convolutional (GCN) layers, fully connected (FC) layers and two-branch convolutional (CNN) layers: GCN layers and FC layers encode the source pose sequence and target view, then CNN branches learn a lower triangular matrix and an upper triangular matrix, respectively, finally they are multiplied to formulate the full-rank transformation matrix.
		For the purpose of adversarial training, we further devise a condition discriminator that distinguishes whether the pose sequence is true or generated.
		To enable the high-level correlation learning, we propose a plug-and-play module, named multi-scale hypergraph convolution (HGC), to replace the spatial graph convolutional layer in baseline, which could simultaneously model the joint-level, part-level and body-level correlations.
		Extensive experiments on two large gait recognition datasets, i.e., CASIA-B and OUMVLP-Pose, demonstrate that our method outperforms the baseline model and existing pose-based methods by a large margin.
	\end{abstract}


	\section{Introduction}
	\IEEEPARstart{I}{n} contrast to some biometric features such as fingerprint and palmprint, gait of human beings could be obtained in a non-contact way, and thus shows the great potential in security surveillance and public safety.
	Gait recognition~\cite{GaitGraph,choi2019skeleton,zou2020deep}, which aims to identify human beings by their continuous walking patterns, has recently received extensive attention from the computer vision community.
	Existing gait recognition methods can be categorized as the silhouette-based methods and model-based methods.
	The former mainly extract the temporal features from binary silhouettes by handcrafted extractor~\cite{GaitSurvey,TCDesc} or deep convolutional neural network (CNN)~\cite{MT3D,ACL,3DLocal,UGaitNet}.
	While the latter first reconstruct human body shape~\cite{tang2016robust} or geometrical structure~\cite{choi2019skeleton,GaitGraph,PoseGait,JRCS} via a shape or pose estimator, then learn features from the reconstructed shape or pose.
	Compared with the silhouette-based methods, the advantage of model-based ones is that they are not sensitive to the illumination and dressing changes.

	\begin{figure}[t]
		\begin{center}
			\includegraphics[width=0.38\textwidth]{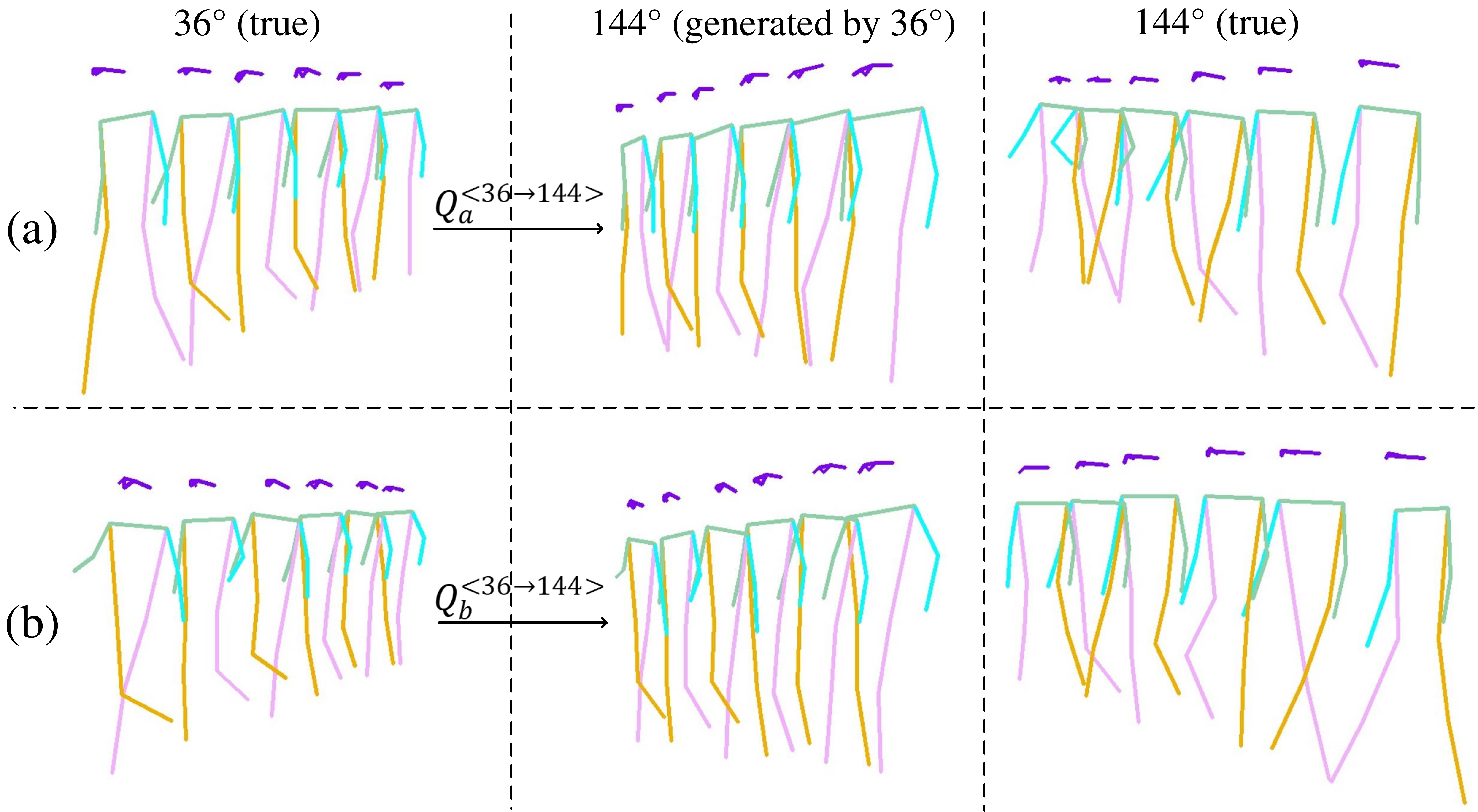}
		\end{center}
		\caption{Column 1 and column 3: the cross-view intra-class variance is much larger than inter-class variance under the same view.
			Column 1 and column 2: this paper proposes to generate multi-view pose sequences for each single-view pose sample via the learning-based linear transformation matrices $Q$, so that each sample could contain complete-view pose sequences.
		}
		\label{fig_view_difference}
	\end{figure}
	
	Among the model-based gait recognition algorithms, the human walking pose, depicted by the motion of skeleton joints, is the most widely-used reconstruction data.
	Considering that estimating 3D human pose from a single image is an ill-posed problem, existing pose-based or skeleton-based methods~\cite{GaitGraph,LGSD,PoseGait} mainly learn gait features from 2D pose.
	Due to the natural graph structure of pose joints, the graph-based methods have become the mainstream technology.
	For example, GaitGraph~\cite{GaitGraph} proposed a strong baseline based on ResGCN~\cite{ResGCN}, which mainly composed of the spatial graph convolutional and temporal convolutional blocks.
	However, the 2D pose would suffer a large intra-class variance when the camera view changes, even though the cross-view intra-class variance would be much larger than the inter-class variance under the same view as shown in Fig.~\ref{fig_view_difference}.
	
	For silhouette-based gait recognition, to reduce the cross-view variance, existing methods~\cite{MGAN,MvGGAN} usually employ the generative adversarial network (GAN) to generate multi-view silhouettes for each single-view silhouette.
	This is achieved by the adversarial training of the silhouette generator and silhouette discriminator.
	However, limited by the representation power of graph generator and GAN model, it is very difficult to generate high-quality temporal pose graph, especially for modeling the large swings of the hands and feet during continuous walking.
	To tackle this problem, this paper proposes a geometry-based multi-view pose generation pipeline, which fully exploits the spatial geometric relationship between cross-view poses.
	According to the principle of camera imaging, we find and derive that the 2D coordinates of cross-view pose graphs satisfy a linear transformation $Q \in R^{3 \times 3}$.
	Concretely, for pose graph $G^{<\alpha>}$ and $G^{<\beta>}$ corresponding to view $\alpha$ and $\beta$, there exists a full-rank matrix $Q^{<\alpha \rightarrow \beta>}$ such that $G^{<\beta>} = Q^{<\alpha \rightarrow \beta>} G^{<\alpha>}$.
	
	Therefore, this paper aims to learn a linear transformation matrix $Q^{<\alpha \rightarrow \beta>}$ from the given source pose sequence $G^{<\alpha>}$ and target view $\beta$.
	To this end, we implement a lower-upper generator (LUGAN) consisting of graph convolution (GCN) layers, fully connected (FC) layers and CNN layers:
	the GCN layers and FC layers first encode the source graph and target view, and the two-branch CNN layers respectively learn a lower triangular matrix $Q^L$ and an upper triangular matrix $Q^U$ from the encoded features, finally the \textbf{full-rank} matrix $Q^{<\alpha \rightarrow \beta>}$ is formulated by multiplying $Q^U$ and $Q^L$.
	This is inspired by the principle that \textit{LU decomposition is possible only for full-rank matrices}.
	To train the generator, we devise a condition discriminator for the adversarial training.
	When training the generator, we hope the pose sequence transformed by $Q$ can be recognized as the true target pose by the discriminator under the condition of the source pose and target view.
	When training the discriminator, it classifies the target pose and transformed pose as true and fake, respectively.

	In addition to neglecting the cross-view variance, another shortcoming of existing methods is that they only learn the joint-level correlation and thus fail to model high-level correlations of human body.
	To resolve this, we propose a differential hypergraph convolution module to replace the GCN layer of baseline model.
	Specifically, we devise a 3-order hypergraph convolution to separately model the joint-level, part-level and body-level correlations, where hyperedge in the higher order hypergraph connects more nodes.
	This hypergraph convolution is integrated as a differential plug-and-play module in framework to learn gait features from the complete-view human pose.
	
	To learn the gait feature from the source pose sequence and generated sequences, this paper proposes an architecture as shown in Fig.~\ref{fig_framework}, which is composed of two branches: a separate branch to learn the source pose feature, and a multi-head branch to learn the generated pose feature.
	Finally we concatenate them together as the final gait representation.
	We test our method on two gait recognition datasets, i.e., CASIA-B~\cite{CASIA-B} and OUMVLP-Pose~\cite{OU-ISIR}.
	The experimental results demonstrate that our method outperforms the baseline model and other existing model-based gait recognition model by a large margin.
	For example, on CL\#1-2, a subset of CASIA-B, the recognition accuracy improves from 65.7\% to 75.4\% after introducing the multi-view pose generation and hypergraph convolution.
	The main contributions of this paper are three-fold:
	\begin{itemize}
		\item To reduce the cross-view gait variance, we propose a geometry-based multi-view pose generation pipeline named LUGAN, where the generator learns a full-rank transformation matrix from the source pose sequence and target view, meanwhile, a condition discriminator is implemented to perform the adversarial training;
		\item To learn the high-level human body correlations, we propose a multi-scale hypergraph convolution module to replace the graph convolution layer, where the part-level and body-level correlations are modeling by hyperedge connecting multiple nodes;
		\item The experimental results verify the validity and generalization of the multi-view pose generation and hypergraph convolution module, in which the former could gain a higher improvement.
	\end{itemize}

	The remainder of this paper is organized as follows: 
	Section~\ref{RW} introduces recent studies related with this paper; 
	Section~\ref{method} elaborates the methodologies;
	Section~\ref{experiment} presents the experiments and visualizations;
	Section~\ref{conclusion} draws brief conclusions.

	\section{Related Works}
	\label{RW}
	
	\subsection{Silhouette-based Gait Recognition}
	\label{SbGR}
	The silhouette-based gait recognition methods learn the human body features from the binary silhouettes using CNN.
	To capture the temporal clues from the gait silhouettes, both the 3D CNN~\cite{MT3D,GLConv,3DLocal} and 2D CNN~\cite{CSTL,ACL} have been employed as the backbone for feature extraction.
	For the 3D CNN, MT3D~\cite{MT3D} proposed a multi-scale 3D network to extract both the small and large temporal scale features;
	GLConv~\cite{GLConv} developed the local and global 3D convolutional layers to obtain the local details and global information;
	3DLocal~\cite{3DLocal} integrated a universal module named 3D local operations  into the backbone to learn the fin-gained feature of each part.
	For the 2D CNN, a crucial step is the temporal aggregation from the feature sequence to the silhouette representation:
	CSTL~\cite{CSTL} proposed an attention-based multi-scale aggregation with fully considering the context information;
	ACL~\cite{ACL} combined the long short-term memory (LSTM)~\cite{LSTM} and attention mechanism to aggregate the silhouette features, moreover, it utilized an angle center loss to reduce the cross-view variance.
	Meanwhile, a lot of studies~\cite{GaitGAN,MGAN,MvGGAN} have explored to reduce the silhouette difference under different views via generative models:
	GaitGAN~\cite{GaitGAN} proposed a GAN model containing two discriminators to generate invariant gait images;
	MGAN~\cite{MGAN} integrated CycleGAN~\cite{CycleGAN} and StarGAN~\cite{StarGAN} to generate multi-view silhouette images for each sample;
	MvGGAN~\cite{MvGGAN} implemented a multi-task generative adversarial network to extract more features from multi-view gait sequences.
	While in this paper, we employ the adversarial training for multi-view pose generation to reduce the intra-class variance.
	
	\subsection{Model-based Gait Recognition}
	\label{MbGR}
	The model-based gait recognition methods first use a shape or pose detector to estimate the human body shape or skeleton structure, then identify human beings from the 3D human shape~\cite{tang2016robust,zhao20063d} and skeleton pose.
	For the 3D shape, Tang \textit{et al.}~\cite{tang2016robust} obtained the human body shape deformation via Laplacian deformation energy function and then utilized a gait partial similarity matching method for gait recognition;
	Zhao \textit{et al.}~\cite{zhao20063d} constructed the 3D human model and achieved the motion tracking via a local optimization algorithm.
	
	Compared to the 3D shape-based methods, the pose-based ones have attracted more attention, and they can be categorized as traditional algorithm-based, CNN-based and GCN-based.
	For the first category, Choi \textit{et al.}~\cite{choi2019skeleton} proposed a two-stage linear matching method for frame-level skeleton matching, where the body symmetry is utilized as the measurement of the skeleton quality; 
	LGSD~\cite{LGSD} proposed multiple local descriptors to represent the graphical skeleton, including the position local pattern, angle local pattern and so on, and learned the skeleton features via a pairwise similarity network.
	For the CNN-based methods, 
	PoseGait~\cite{PoseGait} estimated the 3D pose from the 2D pose and learned the 3D pose feature via deep CNN;
	An \textit{et al.}~\cite{OU-ISIR} built a large-scale skeleton-based gait database named OUMVLP-Pose and presented a CNN-based baseline.
	And for the GCN-based ones, 
	GaitGraph~\cite{GaitGraph} presented another GCN-based pipeline to learn the gait features from the 2D pose, which is adopted as the baseline of this paper;
	JointsGait~\cite{JointsGait} took advantage of the graph convolutional network and joints relationship pyramid mapping layer to extract spatio-temporal gait features.
	However, existing skeleton-based methods did not explicitly deal with the large cross-view variance, and our method could achieve the complete-view gait recognition by multi-view pose generation.

	\subsection{Graph Convolutional Network}
	\label{GCN}
	Generally, the graph convolutional network~\cite{GCN,GCN1,GCN2} refers to the spectral-based graph neural network, which enables the graph representation learning via the graph filtering in the spectral domain. 
	The earliest graph convolutional model~\cite{GCN1} directly learned a graph filter via back propagation algorithm;
	then the ChebyNet~\cite{GCN2} proposed to approximate the graph filter via Chebyshev inequality to reduce the filter parameters;
	GCN~\cite{GCN} achieved the layer-wise graph convolutional structure to alleviate overfitting and increase non-linearity.
	Meanwhile, a great number of GCN variants have been proposed to adapt to different tasks.
	For instance, the hypergraph convolution models~\cite{HGCN1,HGCN2} were raised for high-level correlation learning, in which a hyperedge in hypergraph can connect any number of nodes.
	
	In recent years, the graph convolutional network has shown its privilege in  pedestrian recognition~\cite{ST-GCN,MGH,AAGCN} and skeleton feature learning~\cite{STGCN,Shift-GCN}.
	For the pedestrian recognition, ST-GCN~\cite{ST-GCN} employed the GCN model to learn the part-level correlations for video-based person re-identification (ReID);
	MGH~\cite{MGH} achieved the multi-granular feature learning via hypergraph convolution model;
	AAGCN~\cite{AAGCN} utilized the low-pass filtering property of GCN to smooth the intra-class embeddings, which learned the intra-class adjacency via metric learning and non-linear mapping.
	For the skeleton feature learning, STGCN~\cite{STGCN} first applied GCN in the skeleton-based action recognition, which constructed the edge between spatial and temporal adjacent nodes;
	inspired by shift CNN, ShiftGCN~\cite{Shift-GCN} proposed an efficient GCN model for skeleton-based action recognition, which gained comparable results with less learning-based parameters.
	
	\subsection{Generative Adversarial Network}
	\label{GAN}
	The generative adversarial network~\cite{GAN} is an excellent generative model, which learns the distribution of real data via adversarial training.
	It was first proposed by Goodfellow \textit{et al.}~\cite{GAN}, in which a generator and a discriminator play a \textit{minmax} game to promote each other.
	A large number of variants have been proposed to improve the initial GAN.
	For example, WGAN~\cite{WGAN} employed the Wasserstein divergence to measure the distance between the distribution of real data and that of generated data;
	Conditional GAN~\cite{CGAN,SGAN} enabled the generator to generate samples corresponding to a specific label;
	CycleGAN~\cite{CycleGAN} achieved the pixel-level image translation via a cycle consistency loss.
	
	In pedestrian recognition areas, the GAN models are widely used in image generation for the purpose of smaller intra-class variance.
	For example, PN-GAN~\cite{PN-GAN} proposed a deep person image generation model for image-based ReID, which generated eight canonical pose images for each pedestrian.
	In the infrared-visible person ReID, GAN models could reduce the variance between the infrared domain and visible domain:
	cmGAN~\cite{CMReID1} devised a cutting-edge generative adversarial training strategy to train the cross-modality image generative network;
	cmPIG~\cite{CMReID2} combined CycleGAN~\cite{CycleGAN} to generate the paired-images for cross-modality ReID, so that each pedestrian sample contains a infrared image and a visible one. 
	

	\begin{figure*}[h]
		\begin{center}
			\includegraphics[width=0.98\textwidth]{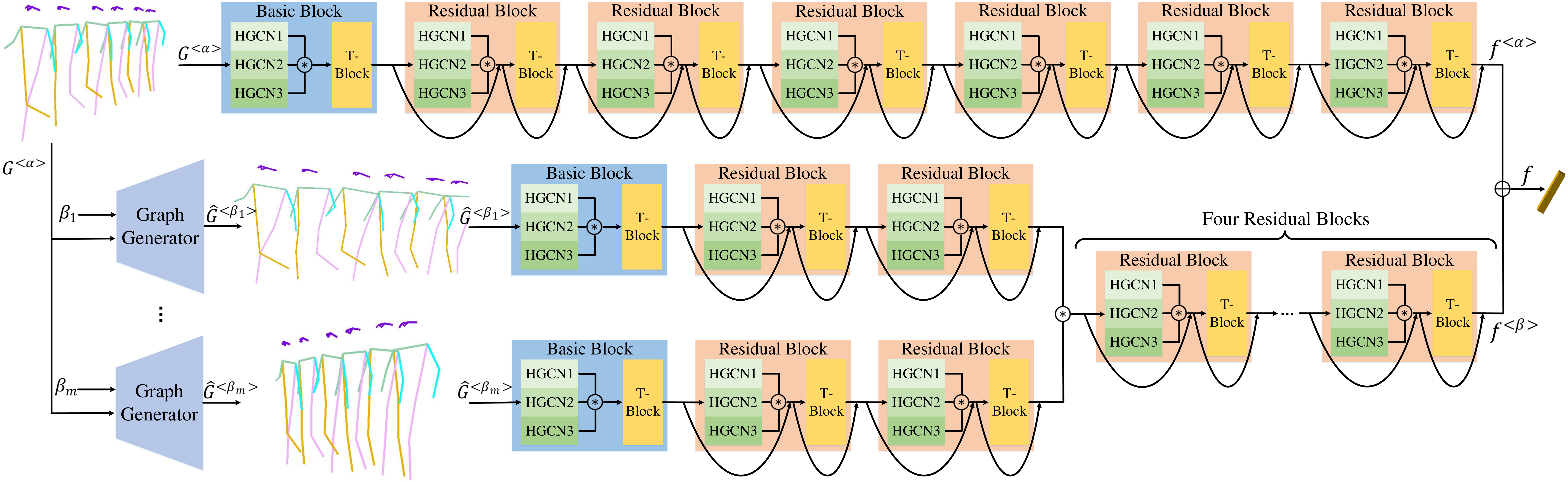}
		\end{center}
		\caption{Schematic of the proposed framework, where $\circledast$ and $\oplus$ represent the average and concatenation operation, respectively.
			The graph generator takes the source pose sequence $G^{<\alpha>}$ and target view $\beta_i$ as input and generates fake pose sequences $\widehat{G}^{<\beta_i>}$.
			Then we devise a two-branch structure to separately learn the source feature $f^{<\alpha>}$ and generative feature $f^{<\beta>}$.
			Each branch consists of multiple basic and residual blocks, and each block is composed of a hypergraph convolution (HGC) and temporal convolution (T-block) operator.
		}
		\label{fig_framework}
	\end{figure*}

	\section{Methodologies}
	\label{method}
	
	\subsection{Preliminaries and Overview}
	\label{PaO}
	Among the model-based gait recognition methods, the human pose is the most widely-used data to represent the walking patterns.
	Specifically, these methods first employ a human pose detection model, e.g., HRNet~\cite{HRNet} or Openpose~\cite{Openpose}, to detect pedestrian pose keypoints, and thus learn the pose features via a neural network.
	In this paper, we depict the human pose by the skeleton graph.
	Concretely, we denote the pose sequence with $T$ frames corresponding to $\alpha^\circ$ camera view as $G^{<\alpha>}=\{G_t | t=1,2,\cdots,T\}$, in which $G_t$ is composed of nodes $V_t$ referring to the pose keypoints and edges $E$ indicating the node connections.
	We assume the number of pose keypoints is $N$, then $V_t$ consists of $v_{t,i}$ for $i=1,2,\cdots,N$.
	We define the initialized feature of $v_{t,i}$ as $(x_{t,i},y_{t,i})$, the keypoint position in the 2D image.
	The edges $E$ are fixed over time, and they could be represented by a adjacency matrix $A \in R^{N \times N}$, where $A_{ij}$ is equal to 1 if $v_{t,i}$ is connected with $v_{t,j}$ and 0 if not.
	
	To reduce the large intra-class variance due to the change of camera view, this paper proposes a pose generation pipeline to generate multiple pose sequences corresponding to different views from a given pose sequence and target views.
	To this end, we exploit the geometric relationships between cross-view pose sequences.
	Specifically, we propose to learn a linear transformation matrix $Q^{<\alpha \rightarrow \beta_i>}$ via adversarial training, which transforms the source graph $G^{<\alpha>}$ to the target graph $\widehat{G}^{<\beta_i>}$.
	As shown in Fig.~\ref{fig_framework}, we take the source graph sequence $G^{<\alpha>}$ and target angle $\beta_i$ as the input of the graph generator, then it would generate the fake pose graph sequence $\widehat{G}^{<\beta_i>}$ corresponding to $\beta_i$.
	The detailed architecture and training strategies of the pose generation pipeline are presented in Section~\ref{MVPG}.
	
	As can be seen in Fig.~\ref{fig_framework}, we adopt the two-branch architecture to learn the pose feature: the top branch learns the source feature $f^{<\alpha>}$ from the input pose $G^{<\alpha>}$, and the bottom branch learns the generative feature $f^{<\beta>}$ from the generated pose $\widehat{G}^{<\beta_i>}$.
	For simplicity, we abbreviate them as the source branch and generative branch, respectively.
	As GaitGraph~\cite{GaitGraph}, the overall structure of each branch in architecture follows the design of ResGCN~\cite{ResGCN} with adaptions.
	Specifically, each branch contains two types of blocks, i.e., the basic block and residual block, where the latter adds extra residual connections compared to the former.
	The source branch is composed of one basic block and six residual blocks, and the output dimensions of each block are presented in Table~\ref{table_out}.
	The generative branch is implemented to be a multi-head structure, where each head corresponds to a specific angle $\beta_i$.
	Each head has its own network weights, and the remaining four residual blocks in the generative branch share weights.
	The output dimension of each block in generative branch is equal to that in source branch.

	\begin{table}[t]
		\centering
		\caption{Output dimension of each block. $T$ and $N$ denote the sequence length and the number of nodes, respectively.}
		\begin{tabular}{ccc}
			\hline
			NO. & Layer          & Output \\ \hline
			1      & Basic block     & $T \times N \times 64$       \\
			2      & Residual block  & $T \times N \times 64$       \\
			3      & Residual block  & $T \times N \times 32$       \\
			4      & Residual block  & $T/2 \times N \times 128$       \\
			5      & Residual block  & $T/2 \times N \times 128$       \\
			6      & Residual block  & $T/4 \times N \times 256$       \\
			7      & Residual block  & $T/4 \times N \times 256$       \\
			8      & Average pooling & $1 \times 256$       \\ \hline
		\end{tabular}
		\label{table_out}
	\end{table}
	
	The biggest difference between our model and ResGCN~\cite{ResGCN} or GaitGraph~\cite{GaitGraph} is that we propose a plug-and-play hypergraph convolution module to replace its spatial graph convolutional layer.
	This plugin consists of multiple hypergraph convolution operations to achieve the multi-scale and high-level pose feature learning.
	We then perform the average pooling on the multi-scale outputs of multiple hypergraph convolutions for the following network propagation.
	The detailed implementation of the hypergraph convolution is presented in Section~\ref{HGCM}.
	By the source branch and generative branch, we could separately learn $f^{<\alpha>}$ and $f^{<\beta>}$; we concatenate them to formulate the final pose feature.
	The loss functions for model training are presented in Section~\ref{MT}.

	\begin{figure*}[t]
		\begin{center}
			\includegraphics[width=0.98\textwidth]{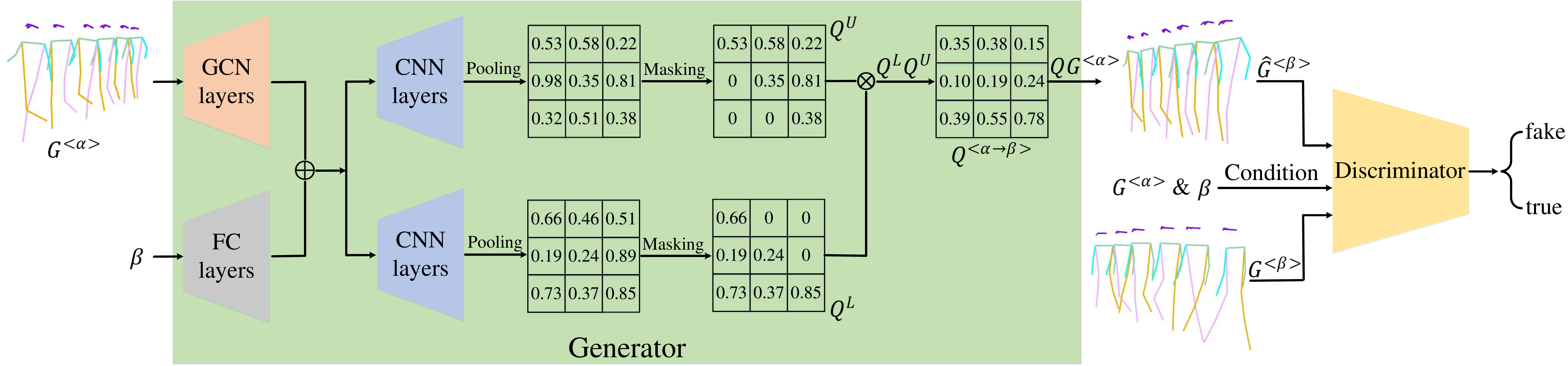}
		\end{center}
		\caption{Schematic of LUGAN. 
			The generator consists of the GCN layers, FC layers and CNN layers.
			It takes the source pose sequence $G^{<\alpha>}$ and target view $\beta$ as input to learn a full-rank transformation matrix $Q^{<\alpha \rightarrow \beta>}$, which transforms $G^{<\alpha>}$ into the target pose $\widehat{G}^{<\beta>}$.
			Under the condition of $G^{<\alpha>}$ and $\beta$, the conditional discriminator distinguishes whether the pose is generated or true.
		}
		\label{fig_gan_pipeline}
	\end{figure*}

	\subsection{Multi-View Pose Generation}
	\label{MVPG}
	Given a pose sequence $G^{<\alpha>}$ corresponding to camera view $\alpha$ and the target angle $\beta$, this paper aims to learn a graph generator to generate the pose $\widehat{G}^{<\beta>}$ to reduce the cross-view variance.
	In this paper, we mainly take advantage of the coordinate transformation in camera imaging for cross-view pose generation.
	
	\textbf{Lemma 1} \emph{Under different camera views, the 2D coordinates of multiple human skeleton keypoints at the same moment share the same cross-view linear transformation $Q$:}
	\begin{equation}
	\begin{gathered}
	\begin{bmatrix} x_{t,i}^{<\beta>} \\ y_{t,i}^{<\beta>} \\ 1 \end{bmatrix} = Q^{<\alpha \rightarrow \beta>} \begin{bmatrix} x_{t,i}^{<\alpha>} \\ y_{t,i}^{<\alpha>} \\ 1 \end{bmatrix}
	\end{gathered},
	\label{eq_linear_transformation}
	\end{equation}
	where ${[x_{t,i}^{<\alpha>}, y_{t,i}^{<\alpha>}]}^T$ represents the 2D position of keypoint $v_{t,i}$ of $G^{<\alpha>}$.
	
	\textbf{Proof.}
	Under the world coordinate system, the 3D coordinates of skeleton point $v_{t,i}$ can be denoted as ${[x_{t,i}^{(w)}, y_{t,i}^{(w)}, z_{t,i}^{(w)}]}^T$.
	Based on the principle of camera imaging, we could derive the following transformation from the world coordinate system to the image coordinate system for camera view $\alpha$:
	\begin{equation}
	\begin{gathered}
	\begin{bmatrix} x_{t,i}^{<\alpha>} \\ y_{t,i}^{<\alpha>} \\ 1 \end{bmatrix} = M_1 M_{2,t}^{<\alpha>} \begin{bmatrix} x_i^{(w)} \\ y_i^{(w)} \\ z_i^{(w)} \\ 1 \end{bmatrix}
	\end{gathered},
	\label{eq_camera_imaging}
	\end{equation}
	and for view $\beta$:
	\begin{equation}
	\begin{gathered}
	\begin{bmatrix} x_{t,i}^{<\beta>} \\ y_{t,i}^{<\beta>} \\ 1 \end{bmatrix} = M_1 M_{2,t}^{<\beta>} \begin{bmatrix} x_i^{(w)} \\ y_i^{(w)} \\ z_i^{(w)} \\ 1 \end{bmatrix}
	\end{gathered},
	\label{eq_camera_imaging_beta}
	\end{equation}
	where $M_1 \in R^{3 \times 3}$ denotes the camera intrinsics matrix, while $M_{2,t}^{<\alpha>}, M_{2,t}^{<\beta>} \in R^{3 \times 4}$ are the camera extrinsics matrices, and their rank is equal to 3.
	The intrinsics matrix represents the intrinsic parameters of the camera, which is a constant matrix;
	the extrinsics matrix consists of the rotation and translation parameters and changes over time.
	
	We abbreviate $M^{<\alpha>} = M_1 M_{2,t}^{<\alpha>} \in R^{3 \times 4}$ and $M^{<\beta>} = M_1 M_{2,t}^{<\beta>} \in R^{3 \times 4}$.
	It's easy to conclude that the rank of both $M^{<\alpha>}$ and $M^{<\beta>}$ is equal to 3.
	Therefore, there exists a full-rank matrix $Q^{<\alpha \rightarrow \beta>} \in R^{3 \times 3}$ to achieve:
	\begin{equation}
	M^{<\beta>} = Q^{<\alpha \rightarrow \beta>} M^{<\alpha>}.
	\label{eq_matrix_transformation}
	\end{equation}
	Based on Eq.~(\ref{eq_matrix_transformation}), Eq.~(\ref{eq_camera_imaging}) and Eq.~(\ref{eq_camera_imaging_beta}), we could derive Eq.~(\ref{eq_linear_transformation}).
	
	In a real scenario, it is very difficult to obtain the accurate frame-by-frame aligned annotation for cross-view pedestrian pose, and there is also no such annotation information in existing datasets.
	Thereby, we cannot directly calculate the linear transformation matrix $Q^{<\alpha \rightarrow \beta>}$ by unaligned $G^{<\alpha>}$ and $G^{<\beta>}$.

	\begin{table}[t]
	\small
	\centering
	\caption{Output dimension of each block in LUGAN. $T$ and $N$ denote the sequence length and the number of nodes, respectively.}
	\begin{tabular}{c|ccc}
		\hline
		Module                                                                    & NO. & Layer          & Output \\ \hline
		\multirow{7}{*}{\begin{tabular}[c]{@{}c@{}}GCN layers\end{tabular}}   & 1      & Basic block     & $T \times N \times 32$       \\
		& 2      & Residual block  & $T \times N \times 32$       \\
		& 3      & Residual block  & $T \times N \times 64$       \\
		& 4      & Residual block  & $T \times N \times 64$       \\
		& 5      & Residual block  & $T \times N \times 64$       \\
		& 6      & Residual block  & $T \times N \times 128$       \\
		& 7      & Residual block & $T \times N \times 128$       \\ \hline
		\multirow{2}{*}{\begin{tabular}[c]{@{}c@{}}FC layers\end{tabular}}   & 1      & FC layer                & $1 \times 64$       \\
		& 2      & FC layer                & $1 \times 128$       \\ \hline
		\multirow{5}{*}{\begin{tabular}[c]{@{}c@{}}CNN layers of $\mathbb{G}$\end{tabular}}   
		& 1      & CNN layer                & $ 64 \times 2N \times 2N $       \\
		& 2      & CNN layer                & $ 128 \times N \times N $       \\
		& 3      & CNN layer                & $ 256 \times N/2 \times N/2 $       \\
		& 4      & CNN layer                & $ 512 \times N/4 \times N/4 $       \\
		& 5      & CNN layer                & $ 1 \times 3 \times 3 $       \\ \hline
		\multirow{6}{*}{\begin{tabular}[c]{@{}c@{}}CNN layers of $\mathbb{D}$\end{tabular}}   
		& 1      & CNN layer                & $ 64 \times 2N \times 2N $       \\
		& 2      & CNN layer                & $ 128 \times N \times N $       \\
		& 3      & CNN layer                & $ 256 \times N/2 \times N/2 $       \\
		& 4      & CNN layer                & $ 512 \times N/4 \times N/4 $       \\
		& 5      & CNN layer                & $ 1 \times 3 \times 3 $       \\ 
		& 6      & Pooling layer                & $ 1 $       \\ \hline
	\end{tabular}
	\label{table_LUGAN_out}
\end{table}

	In this paper, we propose a neural network to learn the transformation matrix $Q$ via adversarial training.
	As mentioned above, $Q$ is a \textbf{full-rank} matrix of size $3 \times 3$.
	To achieve this, we devise a lower-upper (LU) composition pipeline as shown in Fig.~\ref{fig_gan_pipeline} to learn $Q$, i.e., $Q=Q^LQ^U$, in which $Q^L$ and $Q^U$ are the lower triangular matrix and the upper triangular matrix, respectively.
	Specifically, our pipeline takes the source pose sequence $G^{<\alpha>}$ and target view $\beta$ as input, which are separately encoded by the GCN layers and fully-connected (FC) layers.
	Then their respective features are concatenated for two parameter-independent CNN branches to learn $Q^L$ and $Q^U$.
	Thus we can obtain a full-rank matrix $Q$ by matmul product of $Q^L$ and $Q^U$.
	The purpose of above idea is from the principle that \textit{LU decomposition is possible only for full-rank matrices}.
	
	By the learned transformation matrix $Q^{<\alpha \rightarrow \beta>}$ from $G^{<\alpha>}$ and $\beta$, we conduct the linear transformation as Eq.~(\ref{eq_linear_transformation}) on $G^{<\alpha>}$ to generate the fake pose $\widehat{G}^{<\beta>}$:
	\begin{equation}
	\widehat{G}^{<\beta>} = Q^{<\alpha \rightarrow \beta>} G^{<\alpha>},
	\label{eq_transformation_beta}
	\end{equation}
	in which the joint coordinates of $G^{<\alpha>}$ and $\widehat{G}^{<\beta>}$ are normalized to ${[x_{t,i}, y_{t,i}, 1]}^T$.

	\begin{figure*}[ht]
	\centering
	\subfigure[1-Order Hypergraph]{\includegraphics[width=0.3\textwidth]{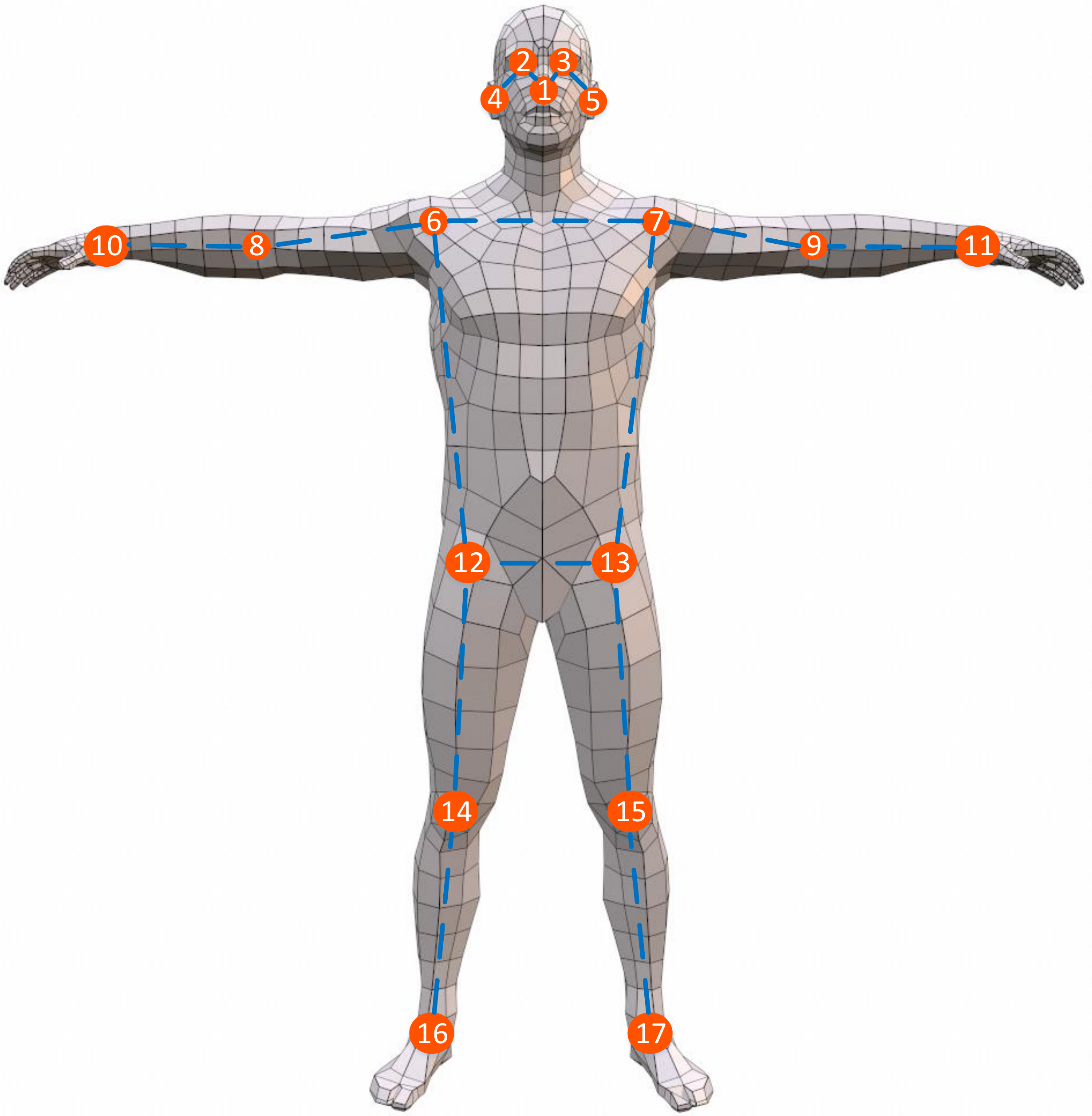}
		\label{fig_HGCN1}}
	\subfigure[2-Order Hypergraph]{\includegraphics[width=0.3\textwidth]{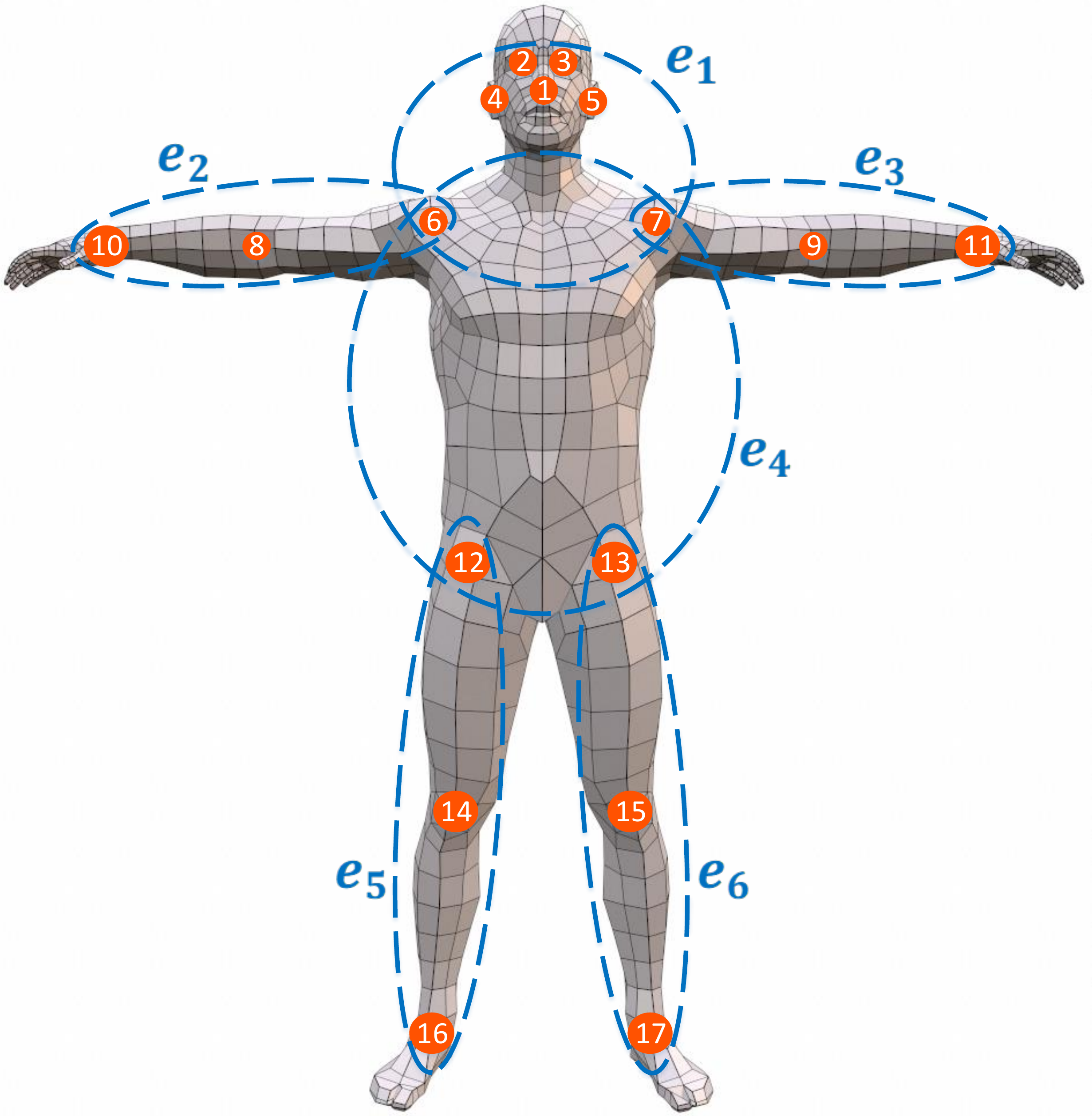}
		\label{fig_HGCN2}}
	\subfigure[3-Order Hypergraph]{\includegraphics[width=0.3\textwidth]{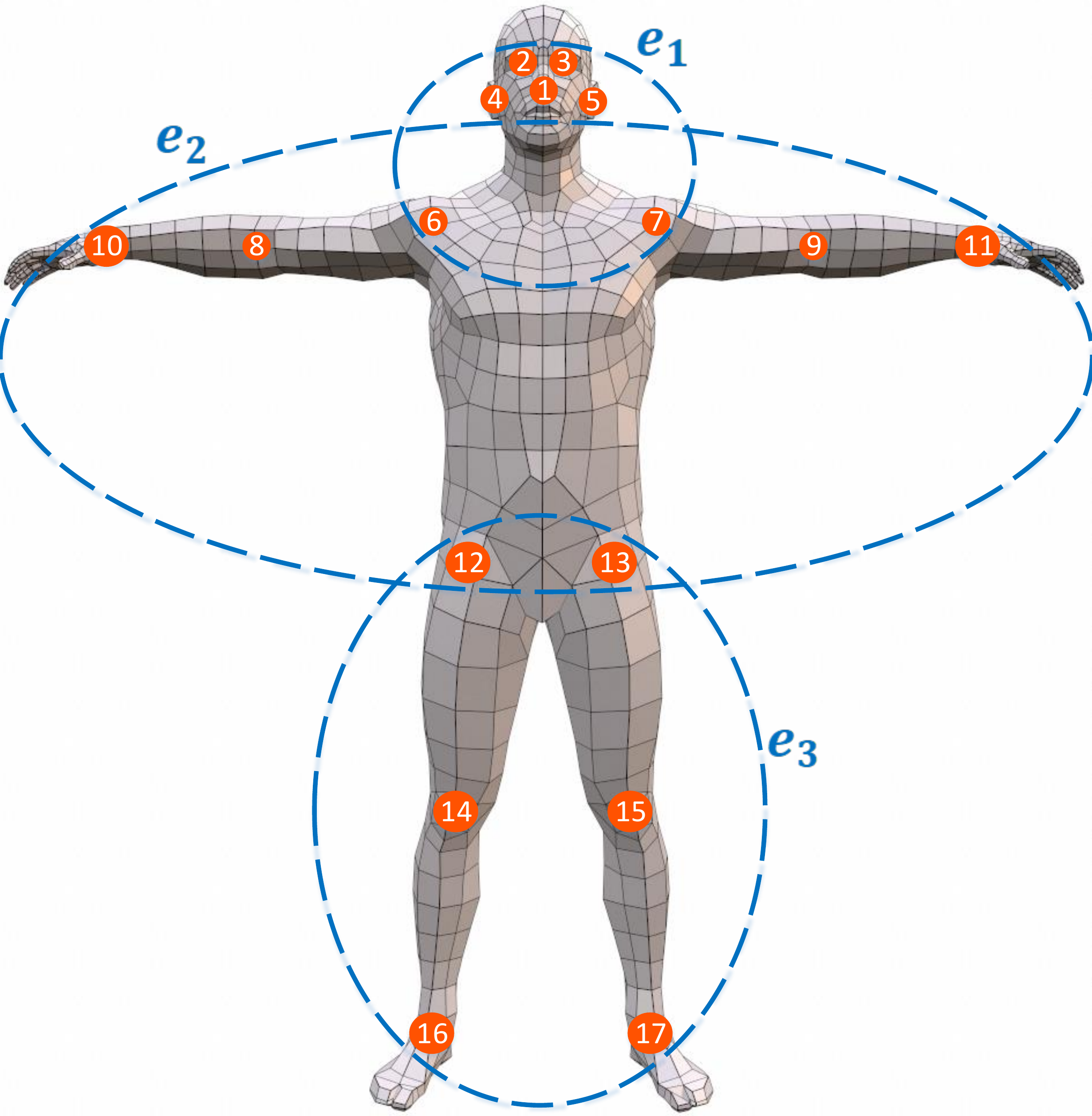}
		\label{fig_HGCN3}}
	\caption{Visualization of the multi-scale hypergraph connections:
		the 1-order hypergraph of (a) degenerates to the normal graph to model the joint-level correlations; the 2-order hypergraph of (b) contains 6 hyperedges to model the part-level correlations; the 3-order hypergraph of (c) contains 3 hyperedges to model the body-level correlations.
	}
	\label{fig_HGCN}
\end{figure*}

	To make the generated pose $\widehat{G}^{<\beta>}$ similar to the true pose $G^{<\beta>}$, we employ the adversarial generating strategy to train our pipeline.
	As shown in Fig.~\ref{fig_gan_pipeline}, we implement a conditional discriminator, whose input denotes the true or generated pose sequence, and condition contains the source pose sequence and target view.
	When training the generator $\mathbb{G}$ (including the GCN layers, FC layers and CNN layers), we hope $\widehat{G}^{<\beta>}$ can be recognized as true by the discriminator $\mathbb{D}$ under the condition $G^{<\alpha>}$ and $\beta$;
	when training $\mathbb{D}$, we fixed $\mathbb{G}$ and classify $\widehat{G}^{<\beta>}$ to be fake.
	The loss function for training $\mathbb{G}$ has the following expression:
	\begin{equation}
	L_{\mathbb{G}} = \Vert I- Q^{<\alpha \rightarrow \beta>}Q^{<\beta \rightarrow \alpha>}\Vert_2 + {(1-\mathbb{D}(\widehat{G}^{<\beta>}|G^{<\alpha>},\beta))}^2,
	\label{eq_training_G}
	\end{equation}
	in which $I$ is the identity matrix to force the invertible transformation.
	And loss function for training $\mathbb{D}$ can be expressed as:
	\begin{equation}
	L_{\mathbb{D}} = {(1-\mathbb{D}(G^{<\beta>}|G^{<\alpha>},\beta))}^2 + {(0-\mathbb{D}(\widehat{G}^{<\beta>}|G^{<\alpha>},\beta))}^2 .
	\label{eq_training_D}
	\end{equation}
	
	\textbf{Architectures.}
	We dub this cross-view pose generation pipeline as LUGAN, and present its detailed network architecture in Table~\ref{table_LUGAN_out}.
	As can be seen, the generator $\mathbb{G}$ contains the following modules: GCN layers consists of seven blocks to encode $G^{<\alpha>}$; FC layers is composed of two fully-connected layers to encode the target view; and CNN layers consists of five convolutional layers to learn $Q^L$ and $Q^U$.
	The conditional discriminator $\mathbb{D}$ has the similar structure with $\mathbb{G}$: 
	GCN layers and FC layers respectively encode the condition $G^{<\alpha>}$ and $\beta$;
	another GCN layers encode the input pose sequence $G^{<\beta>}$ or $\widehat{G}^{<\beta>}$;
	then their features are concatenated for CNN layers to learn the classification logits.
	Compared to $\mathbb{G}$, CNN layers of $\mathbb{D}$ have an extra pooling layer to convert the feature map into classification logits.

	\subsection{Hypergraph Convolution Module}
	\label{HGCM}
	As shown in the framework of Fig.~\ref{fig_framework}, each block (basic block or residual block) contains a multi-scale hypergraph convolution (HGC) layer to learn the multi-level pose correlations.
	In this section, we would elaborate this plug-and-play HGC plugin.
	
	\textbf{Hypergraph convolution revisit.}
	As a generalization of the graph data, the hypergraph could model high-level relations by each hyperedge connecting multiple nodes.
	For example, edge $e_1$ in Fig.~\ref{fig_HGCN3} connects nodes 1, 2, 3, 4, 5, 6 and 7.
	Given a hypergraph, we could define an incidence matrix $H \in R^{N \times M}$, where $N$ and $M$ denote the number of nodes and hyperedges, respectively.
	We define that $H_{i \epsilon}$ is equal to 1 if node $v_i$ is connected by edge $e_{\epsilon}$ and equal to 0 if not.
	For instance, the size of the hypergraph incidence matrix shown in Fig.~\ref{fig_HGCN3} is $17 \times 3$.
	Then the network propagation can be denoted as follows:
	\begin{equation}
	X^{(l+1)} = \sigma (D^{-1/2} H B^{-1} H^T D^{-1/2} X^{(l)} W^{(l+1)}) ,
	\label{eq_HGCN}
	\end{equation}
	where $X^{(l)}$ denotes the node features in the $l$-th layer; $\sigma(\cdot)$ refers to the {\rm ReLU} activation function; $D_{ii} = \sum_{\epsilon=1}^M H_{i \epsilon}$ and $B_{\epsilon \epsilon} = \sum_{i=1}^N H_{i \varepsilon}$ are diagonal matrices; and $W^{(l+1)}$ is a learning-based parameter.
	We note the hypergraph convolution would degenerate to the graph convolution when the hyperedges degenerate to normal edges (each edge connect two nodes).
	For more details, we recommend the readers to refer to the original papers~\cite{HGCN1,HGCN2}.
	
	To achieve the multi-level pose feature learning, we construct 3 hypergraphs as shown in Fig.~\ref{fig_HGCN} to separately model the joint-level, part-level and body-level correlations.
	As can be seen, the 1-order hypergraph degenerates to the normal graph, while the 2-order and 3-order hypergraphs contain 6 and 3 hyperedges, respectively.
	For these three hypergraphs, we denote their incidence matrices as $H_1$, $H_2$ and $H_3$, respectively.
	Then the output of $j$-th hypergraph can be computed by the following formula for $j=\{1,2,3\}$:
	\begin{equation}
	X^{(l+1)}_j = \sigma (D^{-1/2}_j H_j B^{-1}_j H^T_j D^{-1/2}_j X^{(l)} W^{(l+1)}_j) .
	\label{eq_HGCNj}
	\end{equation}
	Next, we perform the average pooling on the multi-head outputs to obtain the feature of $(l+1)$-th layer:
	\begin{equation}
	X^{(l+1)} = \sum_{j=1}^{3} X^{(l+1)}_j .
	\label{eq_mean}
	\end{equation}
	
	In Eq.~(\ref{eq_HGCNj}) and Eq.~(\ref{eq_mean}), three hypergraph convolutions share the same input feature $X^{(l)}$, and all the operations in them are differentiable.
	Thus, this multi-order hypergraph convolution can be utilized as a plug-and-play plugin to replace the spatial graph convolution layer of original ResGCN for multi-level pose correlation learning.

	\subsection{Model Training}
	\label{MT}
	Followed the baseline model GaitGraph~\cite{GaitGraph}, we train our model using the supervised contrastive loss~\cite{SCL}.
	The framework in Fig.~\ref{fig_framework} shows that we can learn the feature $f^{<\alpha>}_i$ from the source pose sequence and feature $f^{<\beta>}_i$ from the generated pose sequence.
	We assume that the corresponding label of $f^{<\alpha>}_i$ and $f^{<\beta>}_i$ is $y_i$.
	We then define the following losses to train our model:
	\begin{equation}
	L_{SCL}^{<\alpha>} = - \frac{1}{n} \sum_{i=1}^{n} \sum_{y_i=y_j} \log \frac{\exp (f^{<\alpha>}_i \cdot f^{<\alpha>}_j / \tau)}{\sum_{y_i \neq y_k} \exp (f^{<\alpha>}_i \cdot f^{<\alpha>}_k / \tau)} ,
	\label{eq_loss_alpha}
	\end{equation}
	\begin{equation}
	L_{SCL}^{<\beta>} = - \frac{1}{n} \sum_{i=1}^{n} \sum_{y_i=y_j} \log \frac{\exp (f^{<\beta>}_i \cdot f^{<\beta>}_j / \tau)}{\sum_{y_i \neq y_k} \exp (f^{<\beta>}_i \cdot f^{<\beta>}_k / \tau)} ,
	\label{eq_loss_beta}
	\end{equation}
	where $n$ denotes the training batch size, and $\tau$ is a hyper-parameter.
	We define the final loss $L$ as the sum of $L_{SCL}^{<\alpha>}$ and $L_{SCL}^{<\beta>}$:
	\begin{equation}
	L = L_{SCL}^{<\alpha>} + L_{SCL}^{<\beta>} .
	\label{eq_loss_final}
	\end{equation}
	When minimizing $L$, features corresponding to the same identity, i.e., $f^{<\alpha>}_i$ and $f^{<\alpha>}_j$, are encouraged to be closed to each other, and features corresponding to different identities are pushed away from each other.

	\section{Experiments}
	\label{experiment}
	
	\subsection{Datasets and Implementations}
	\label{DaI}
	\textbf{Datasets.} To validate our method, we conduct experiments on two gait recognition datasets: CASIA-B~\cite{CASIA-B} and OUMVLP-Pose~\cite{OU-ISIR}.
	CASIA-B consists of 124 identities with three walking statuses: normal walking (NM), wearing a coat (CL) and carrying a bag (BG).
	Each identity in CASIA-B contains 110 sequences captured from 11 camera views: \{0\textdegree, 18\textdegree, 36\textdegree, 54\textdegree, 72\textdegree, 90\textdegree, 108\textdegree, 126\textdegree, 144\textdegree, 162\textdegree, 180\textdegree \}.
	These 124 identities are divided into the training set and testing set according to the ratio of 74:50.
	In the testing set, part of identities in NM subset are taken as the gallery and the rest for probe.
	OUMVLP-Pose~\cite{OU-ISIR} is currently the largest pose database for gait recognition, which contains over 10,000 subjects from 14 camera views: \{0\textdegree, 15\textdegree, 30\textdegree, 45\textdegree, 60\textdegree, 75\textdegree, 90\textdegree, 180\textdegree, 195\textdegree, 210\textdegree, 225\textdegree, 240\textdegree, 255\textdegree, 270\textdegree \}.
	The seqence length in OUMVLP-Pose is in the range from 18 to 35, with an average of 25.
	The training set and testing set are composed of 5,153 and 5,154 subjects, respectively.
	
	\textbf{Implementations.} 
	For the multi-view pose generation pipeline of Fig.~\ref{fig_gan_pipeline}, we train the LUGAN model 20 epochs on CASIA-B and OUMVLP-Pose; we employ Adam~\cite{Adam} optimizer to train models; the learning rate of both the generators and discriminator is set to 0.0001; the features of the source pose sequence $G^{<\alpha>}$ and target angle $\beta$ are both encoded to 128-dimensional.
	In each iteration, we train the generator 50 times and train the discriminator once.
	On CASIA-B dataset, we generate 10 pose sequences for each given single-view sequence, so that each sample contains 11 sequences corresponding to complete camera views.
	While on OUMVLP-Pose, we generate 13 pose sequences for each sample.
	
	For the gait recognition network of Fig.~\ref{fig_framework}, we mostly follow the experimental settings of our baseline model GaitGraph~\cite{GaitGraph} for fair comparison: we perform the random crop on each sample to fix its length;
	we add the Gaussian noise to the original node feature; we train the network 500 epochs with Adam~\cite{Adam} optimizer, whose learning rate is set to 0.001; 
	the dimension of learned gait pose feature is set to 256.

\subsection{Ablation Studies}
\label{AS}
\subsubsection{Comparison with baseline model}
\label{CwB}
\begin{figure}[h]
	\begin{center}
		\includegraphics[width=0.42\textwidth]{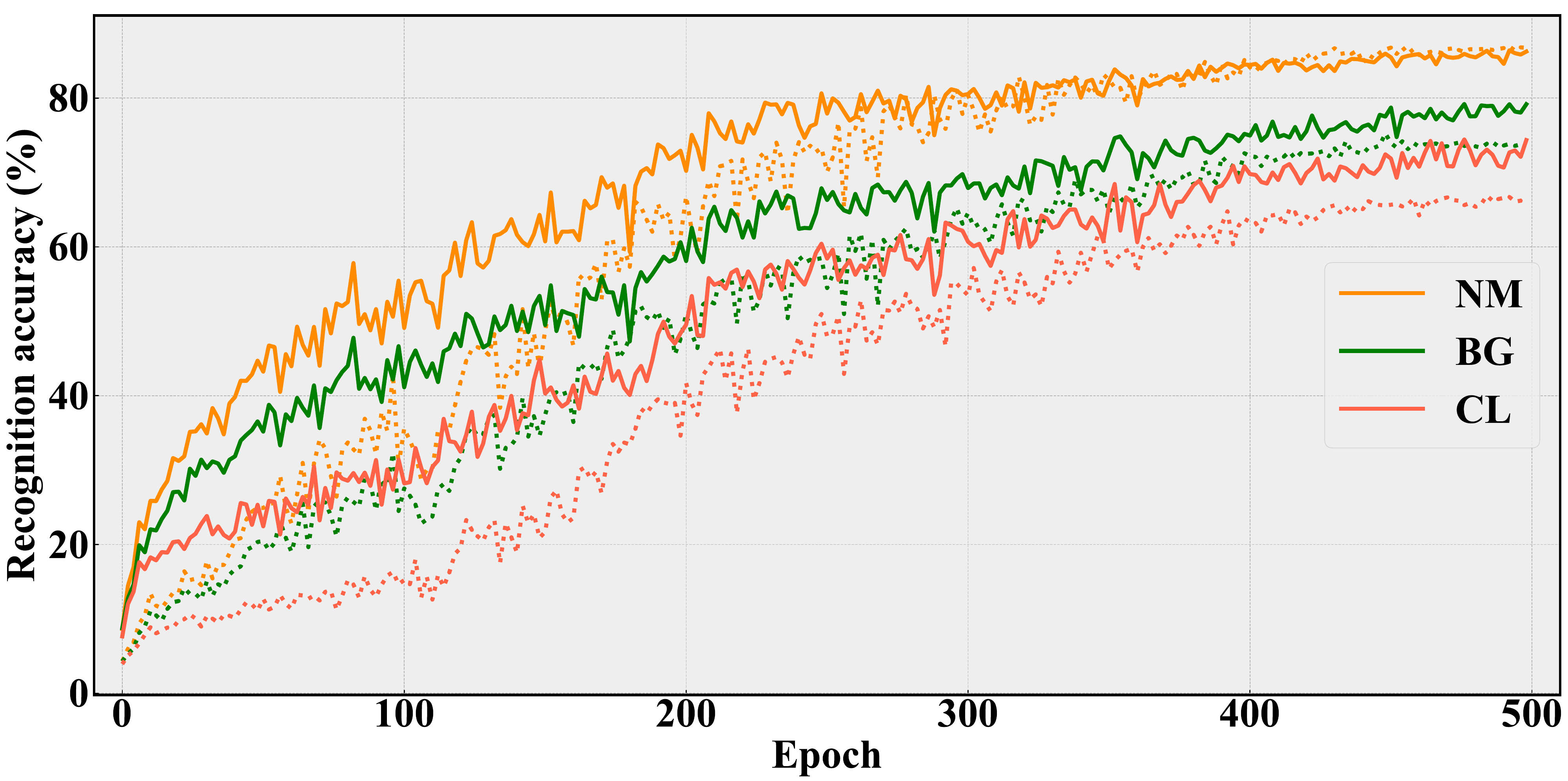}
	\end{center}
	\caption{Visualization of recognition accuracy during training CASIA-B~\cite{CASIA-B}, where the solid line and dashed line denote the performance of LUGAN and baseline, respectively.
	}
	\label{fig_accuracy}
\end{figure}

In this section, we first present the overall comparisons with the baseline model, GaitGraph~\cite{GaitGraph}.
The results on CASIA-B~\cite{CASIA-B} and OUMVLP-Pose~\cite{OU-ISIR} are reported in Table~\ref{table_compare_baseline_casia} and Table~\ref{table_compare_baseline_oumvlp}, respectively.
In these tables, ``LUGAN'' means that we conduct the multi-view pose generation via LUGAN on the basis of baseline, and the basic spatial block in backbone is graph convolution (GC) rather than HGC.
While ``LUGAN-HGC'' indicates that we learn gait representation via HGC-based ResGCN from the complete-view pose sequence.
We could draw the following conclusions:
1) the multi-view pose generation by LUGAN could gain a significant improvement compared with the baseline model, i.e., 9.7\% on CL subset of CASIA-B and 5.2\% on OUMVLP-Pose;
2) the HGC-based model outperforms the GC-based, which demonstrates the superiority of the high-level correlation learning.
Moreover, we visualize the recognition accuracy during training in Fig.~\ref{fig_accuracy}.
We conclude that our model converges faster than the baseline.

\begin{table*}[ht]
	\small
	\centering
	\caption{Comparison with the baseline model on CASIA-B~\cite{CASIA-B}. Numbers shown in the table are percentages.}
	\begin{tabular}{cc|ccccccccccc|c}
		\hline
		\multicolumn{2}{c|}{Gallery NM\#1-4}                               & \multicolumn{11}{c|}{0°-180°}                                              & \multirow{2}{*}{mean} \\ \cline{1-13}
		\multicolumn{2}{c|}{Probe}                                         & 0°   & 18°  & 36°  & 54°  & 72°  & 90° & 108° & 126° & 144° & 162° & 180° &                       \\ \hline
		\multicolumn{1}{c|}{\multirow{3}{*}{NM\#5-6}} & Baseline          & 83.9 & 88.1 & 88.3 & 89.6 & 86.7 & 87.4	 & 86.7 & 87.4 & 88.2 & 88.4 & 82.3 & 87.0                  \\
		\multicolumn{1}{c|}{}                         & LUGAN      & 86.8     & 88.0     & 88.2     & 88.6     & 87.9     & 87.0     & 87.7     & 87.3     & 88.4     & 87.5     & 83.6     & 87.1                      \\
		\multicolumn{1}{c|}{}                         & LUGAN-HGC & 89.3     & 88.1     & 89.0     & 89.9     & 87.4     & 88.7     & 87.4     & 88.8     & 88.8     & 87.0     & 87.0     & 88.3                      \\ \hline
		\multicolumn{1}{c|}{\multirow{3}{*}{BG\#1-2}} & Baseline          & 78.0 & 77.9 & 78.1 & 75.5 & 72.8 & 70.1 & 70.6 & 72.4 & 74.5 & 78.0 & 68.7 & 74.2                  \\
		\multicolumn{1}{c|}{}                         & LUGAN      & 78.6     & 77.3     & 80.4     & 80.8     & 76.7     & 75.6     & 78.7     & 80.4     & 81.0     & 82.0     & 70.5     & 78.4                     \\
		\multicolumn{1}{c|}{}                         & LUGAN-HGC & 79.4     & 79.5     & 81.6     & 82.4     & 78.1     & 76.2     & 78.7     & 82.0     & 81.6     & 83.0     & 73.6     & 79.7                      \\ \hline
		\multicolumn{1}{c|}{\multirow{3}{*}{CL\#1-2}} & Baseline          & 62.0 & 60.8 & 64.3 & 65.8 & 67.8 & 65.7 & 70.7 & 63.8 & 67.5 & 68.7 & 65.0 & 65.7                  \\
		\multicolumn{1}{c|}{}                         & LUGAN      & 73.5     & 71.0     & 71.2    & 74.0     & 75.5     & 75.0     & 80.4     & 75.8     & 74.1     & 73.5     & 67.6     & 73.8                      \\
		\multicolumn{1}{c|}{}                         & LUGAN-HGC & 72.8     & 72.3     & 69.4     & 75.2     & 77.0     & 79.6     & 80.5     & 78.1     & 76.3     & 74.9     & 72.8     & 75.4                      \\ \hline
	\end{tabular}
	\label{table_compare_baseline_casia}
\end{table*}

\begin{table*}[ht]
	\small
	\centering
	\caption{Comparison with the baseline model on OUMVLP-Pose~\cite{OU-ISIR}.}
	\begin{tabular}{c|ccccccc|ccccccc|c}
		\hline
		\multirow{2}{*}{Method} & \multicolumn{7}{c|}{0°-90°}                    & \multicolumn{7}{c|}{180°-270°}                 & \multirow{2}{*}{mean} \\ \cline{2-15}
		& 0°   & 15°  & 30°  & 45°  & 60°  & 75°  & 90° & 180° & 195° & 210° & 225° & 240° & 255° & 270° &                       \\ \hline
		Baseline                & 36.6 & 45.5 & 47.9 & 48.4 & 47.5 & 48.9 & 41.8 & 44.6 & 40.3 & 40.1 & 40.5 & 40.9 & 41.7 & 33.3 & 42.7                      \\
		LUGAN                  & 41.5 & 47.1 & 50.4 & 51.3 & 50.4 & 51.0 & 46.4 & 46.8 & 45.5 & 46.7 & 46.8 & 45.8 & 46.2 & 40.1 & 46.8                      \\
		LUGAN-HGC               & 42.6 & 47.7 & 52.1 & 53.6 & 50.4 & 51.6 & 48.0 & 49.6 & 45.8 & 47.5 & 47.6 & 47.0 & 47.4 & 40.7 & 47.9                      \\ \hline
	\end{tabular}
\label{table_compare_baseline_oumvlp}
\end{table*}

In the following sections, we would separately validate the multi-view pose generation pipeline and the HGC-based multi-level correlation learning.
For the former, we first compare our geometry-based multi-view pose generation with 1) 3D pose and 2) adversarial graph generator that learns multi-view pose sequence via a pose encoder and a decoder. 
We then validate the generalization of our method: we perform the cross-dataset multi-view pose generation for each dataset, i.e., generating multi-view pose of OUMVLP-Pose~\cite{OU-ISIR} for each sample in CASIA-B~\cite{CASIA-B}.
Next we verify the validity of lower-upper composition way for learning the full-rank transformation matrix.
For the HGC module, we test the performance of our model under the setting of 1-order, 2-order and 3-order HGC, respectively.
Finally, we validate the number of shared blocks in the generative branch of our architecture.
For simplicity, we will only report the mean rank1 accuracy.

\begin{figure}[t]
	\begin{center}
		\includegraphics[width=0.48\textwidth]{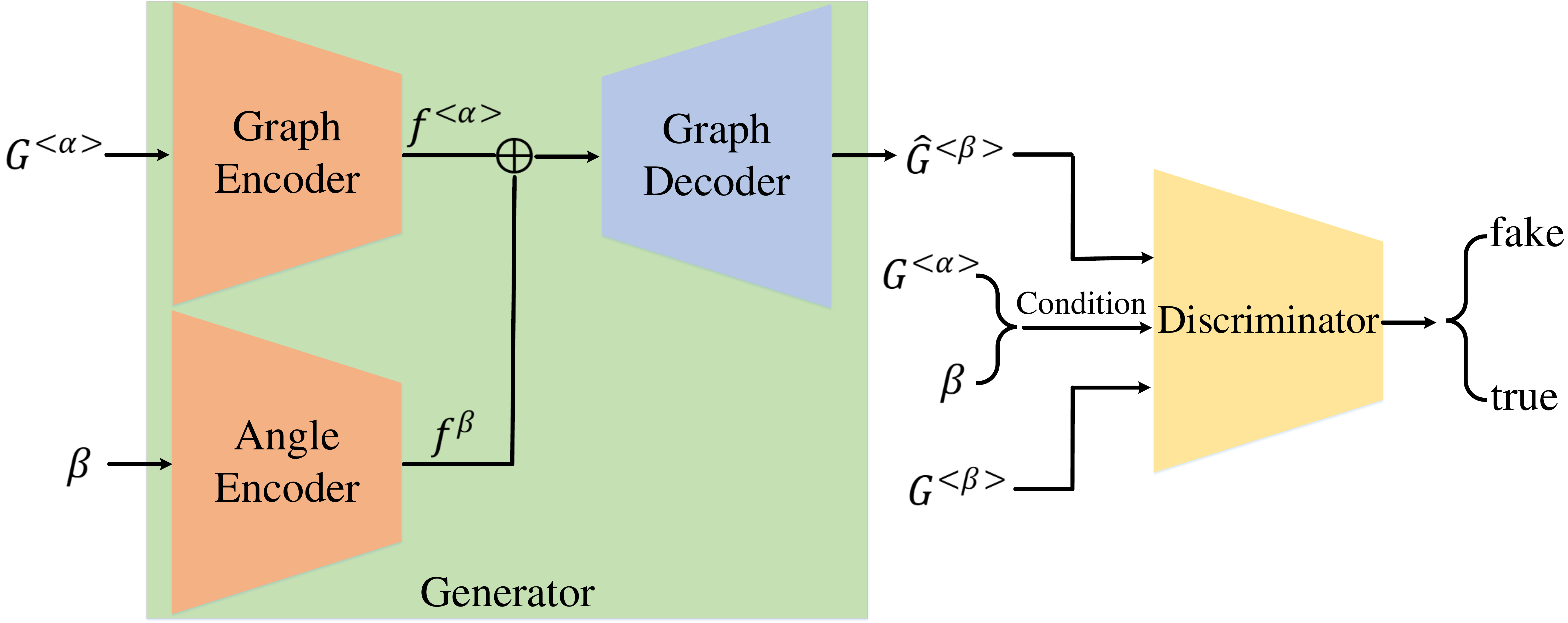}
	\end{center}
	\caption{Pose generator via EDGAN. 
	The generator consists of a graph encoder, an angle encoder	and a pose decoder, while the condition discriminator has the same structure with LUGAN.
}
	\label{fig_EDGAN}
\end{figure}

\begin{table}[t]
	\small
	\centering
	\caption{Comparison with the 3D pose and E\&D generator.}
	\begin{tabular}{c|ccc|c}
		\hline
		\multirow{2}{*}{Methods} & \multicolumn{3}{c|}{CASIA-B} & \multirow{2}{*}{OUMVLP-Pose} \\ \cline{2-4}
		& NM\#5-6  & BG\#1-2 & CL\#1-2 &                         \\ \hline
		Baseline                 & 87.0     & 74.2    & 65.7    & 42.7                    \\
		Baseline+3D              & 85.8     & 73.0    & 58.4    & 37.8                    \\
		EDGAN                    & 87.3     & 77.4    & 67.2    & 46.0                    \\
		LUGAN                    & 87.1     & 78.4    & 73.8    & 46.8                    \\ \hline
	\end{tabular}
	\label{table_com_3d}
\end{table}

\begin{figure*}[t]
	\begin{center}
		\includegraphics[width=0.98\textwidth]{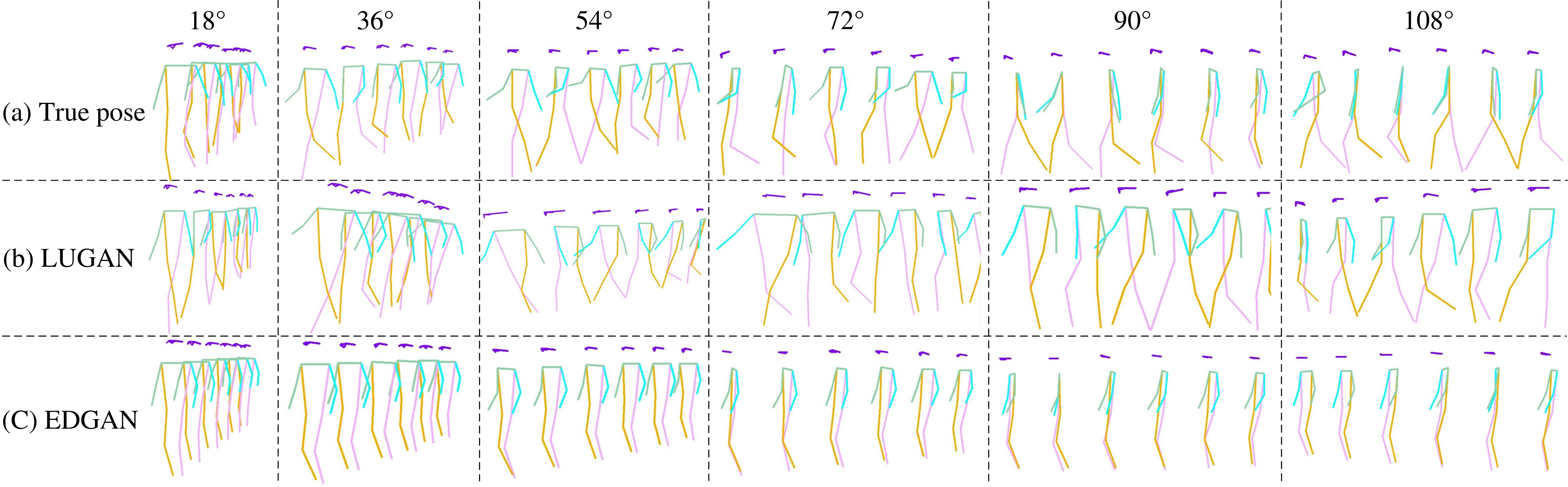}
	\end{center}
	\caption{Visualization of (a) true pose, (b) pose sequence generated by LUGAN and (c) pose sequence generated by EDGAN. As can be seen, EDGAN can not learn the large swings of the hands and feet during continuous walking, in which the pedestrian is sliding rather than walking.}
	\label{fig_vis_pose}
\end{figure*}

\begin{figure*}[t]
	\centering
		\subfigure[HGCN1]{\includegraphics[width=0.3\textwidth]{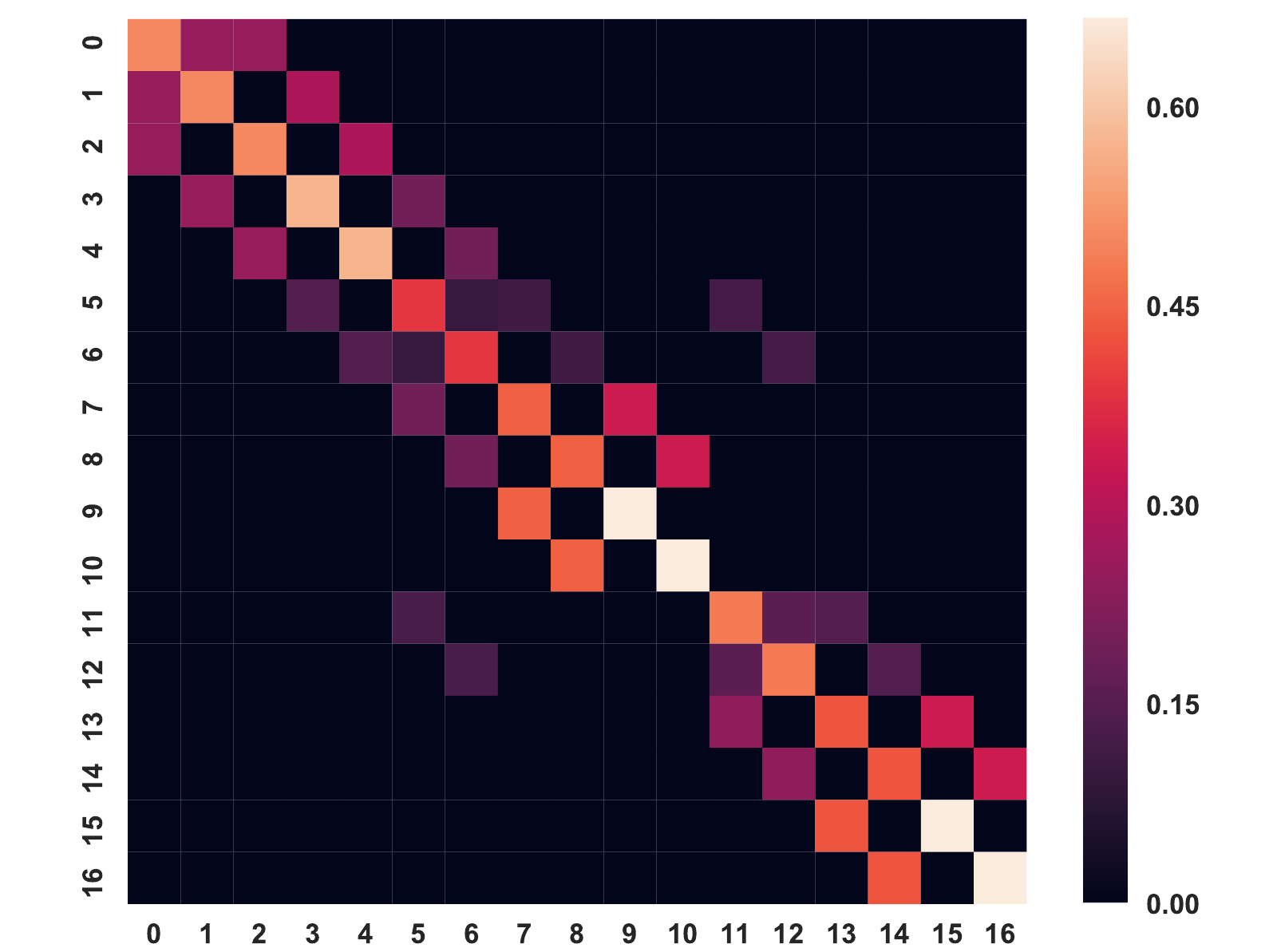}}
	\subfigure[HGCN2]{\includegraphics[width=0.3\textwidth]{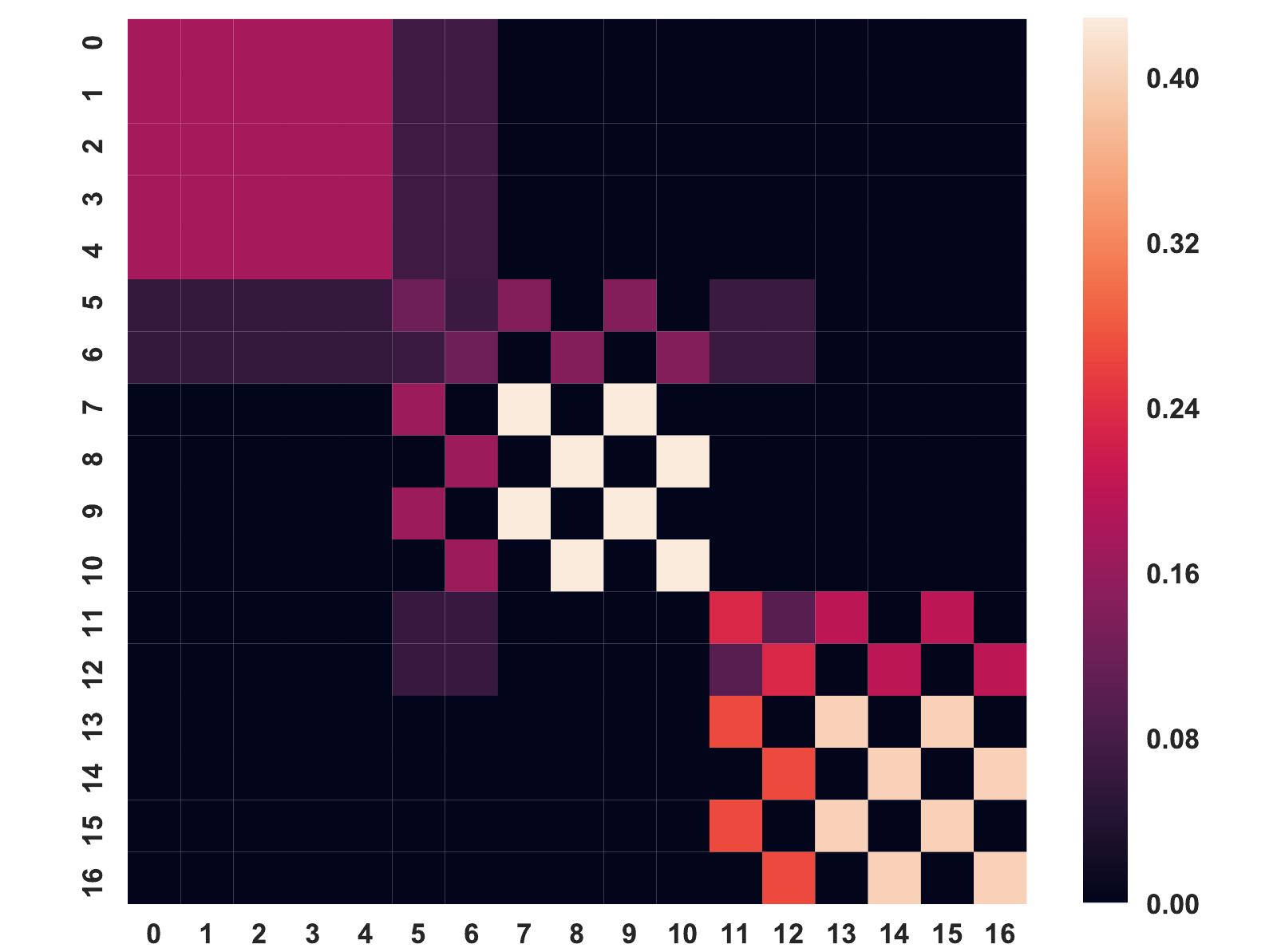}}
	\subfigure[HGCN3]{\includegraphics[width=0.3\textwidth]{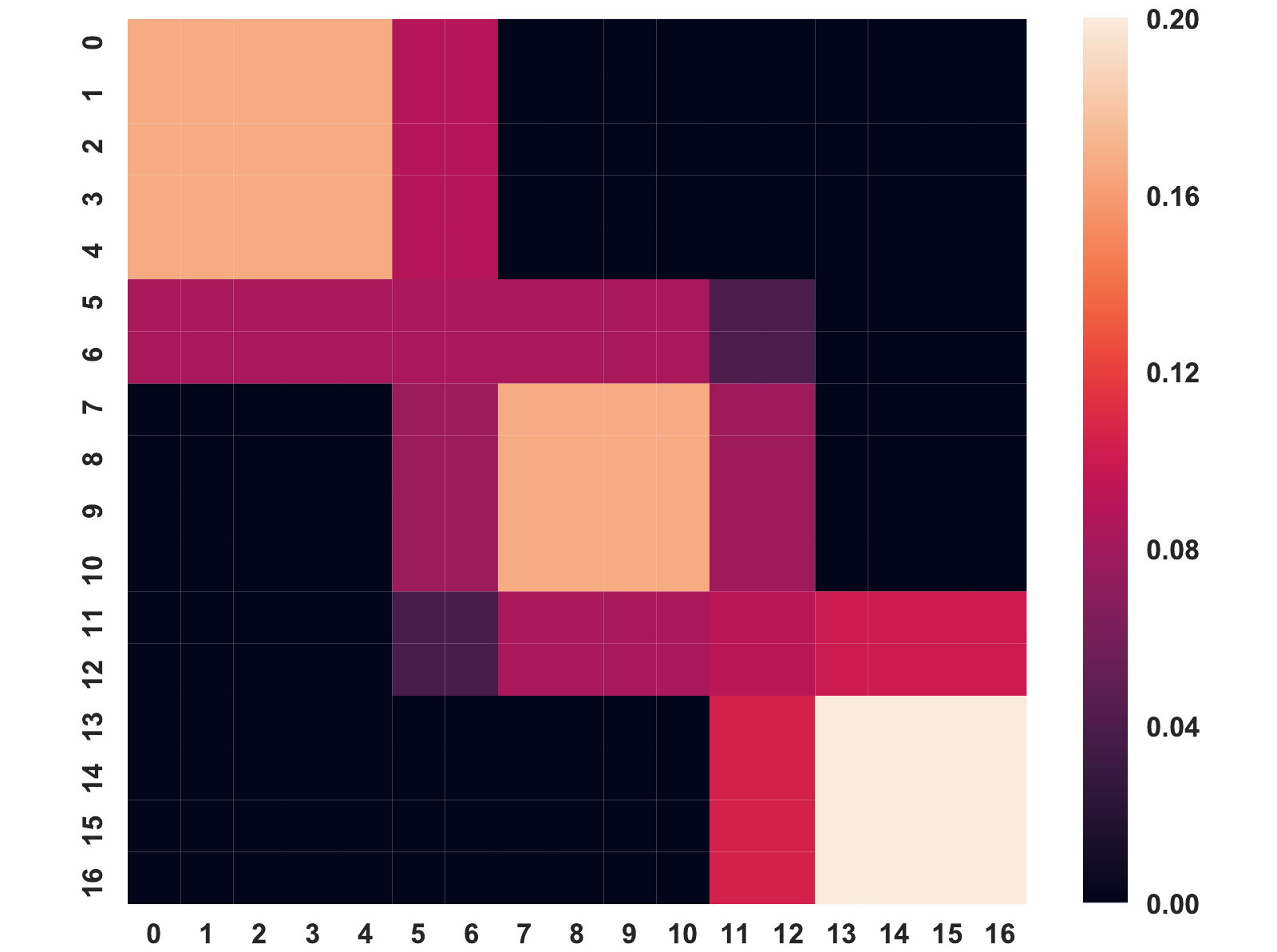}}
	\caption{Visualization of the normalized adjacency $D^{-1/2} H B^{-1} H^T D^{-1/2}$ of each hypergraph.}
	\label{fig_adj}
\end{figure*}

\subsubsection{Comparison with 3D pose and other multi-view pose generation pipeline}
\label{C3PVG}
In this section, we compare LUGAN with 1) 3D pose and 2) the pose generator via encoder-decoder (EDGAN) paradigm as shown in Fig.~\ref{fig_EDGAN}.
For the 3D pose, we employ GAST-Net~\cite{GAST-Net} to evaluate 3D pose from 2D pose and then learn 3D pose features via baseline model.
For the latter, its generator consists of a graph encoder, an angle encoder and a pose decoder, which contain three GCN blocks, two FC layers and four GCN blocks, respectively.
The graph encoder and angle encoder first learn features from $G^{<\alpha>}$ and $\beta$, then the pose decoder directly learns pose coordinates from the encoded features.
And the condition discriminator has the same structure with LUGAN for adversarial training.

We report the results in Table~\ref{table_com_3d}.
As can be seen, the performance of 3D pose is even worse than the single-view 2D pose, which proves that it's difficult to evaluate accurate 3D pose from a single image.
Meanwhile, our LUGAN could outperform EDGAN on both CASIA-B~\cite{CASIA-B} and OUMVLP-Pose~\cite{OU-ISIR}.
In Fig.~\ref{fig_vis_pose}, we visualize (a) true pose samples from CASIA-B, (b) pose sequences generated by LUGAN and (b) pose sequences generated by EDGAN.
The pose sequences generated by LUGAN is similar with the true pose samples, but EDGAN can not model the large swings of the hands and feet during continuous walking.

\begin{table}[t]
	\small
	\centering
	\caption{Validation of the cross-dataset multi-view pose generation. CD-LUGAN denotes that we perform the cross-dataset multi-view pose generation by LUGAN.}
	\begin{tabular}{c|ccc|c}
		\hline
		\multirow{2}{*}{Methods} & \multicolumn{3}{c|}{CASIA-B} & \multirow{2}{*}{OUMVLP-Pose} \\ \cline{2-4}
		& NM\#5-6  & BG\#1-2 & CL\#1-2 &                         \\ \hline
		Baseline                 & 87.0     & 74.2    & 65.7    & 42.7                    \\
		CD-LUGAN                    & 86.8     & 76.5    & 68.7    & 43.4                    \\ \hline
	\end{tabular}
	\label{table_cdlugan}
\end{table}

\subsubsection{Validation of generalization}
\label{VoG}
Limited by the training samples in each dataset, the proposed LUGAN could only generate pose sequences corresponding to existing camera views.
In this section, we verify the generalization of LUGAN by the cross-dataset multi-view pose generation.
Specifically, for each pose sample on CASIA-B~\cite{CASIA-B}, we use the LUGAN model trained on OUMVLP-Pose~\cite{OU-ISIR} to generate 14 pose sequences corresponding to view \{0\textdegree, 15\textdegree, 30\textdegree, 45\textdegree, 60\textdegree, 75\textdegree, 90\textdegree, 180\textdegree, 195\textdegree, 210\textdegree, 225\textdegree, 240\textdegree, 255\textdegree, 270\textdegree \}, so that each sample contains 15 sequences.
And for pose sample on OUMVLP-Pose, we generate 11 pose sequences corresponding to 11 views of CASIA-B, thereby each sample contains 12 sequences.

As shown in Table~\ref{table_cdlugan}, even though the improvement is smaller than inner-dataset multi-view pose generation, the cross-dataset generation still outperforms the baseline model.
This benefits from the multi-view information compensation, and also demonstrates the generalization of our LUGAN.

\begin{table}[t]
	\small
	\centering
	\caption{Validation of the cross-dataset multi-view pose generation. QGAN denotes that we directly learn a transformation matrix $Q \in R^{3 \times 3}$ instead of $Q^L$ and $Q^U$.}
	\begin{tabular}{c|ccc|c}
		\hline
		\multirow{2}{*}{Methods} & \multicolumn{3}{c|}{CASIA-B} & \multirow{2}{*}{OUMVLP-Pose} \\ \cline{2-4}
		& NM\#5-6  & BG\#1-2 & CL\#1-2 &                         \\ \hline
		Baseline                 & 87.0     & 74.2    & 65.7    & 42.7                    \\
		QGAN                    & 87.0     & 78.0    & 72.3    & 46.3                    \\
		LUGAN                    & 87.1     & 78.4    & 73.8    & 46.8                    \\ \hline
	\end{tabular}
	\label{table_QGAN}
\end{table}

\subsubsection{Impact of LU multiplication}
\label{IoLM}
The LUGAN pipeline of Fig.~\ref{fig_gan_pipeline} consists of two CNN branches to separately learn a lower matrix $Q^L$ and an upper matrix $Q^U$ to ensure that the transformation matrix $Q$ can be full-rank.
In this section, we compare LUGAN and QGAN model that directly learns a transformation matrix $Q \in R^{3 \times 3}$, in which the latter only contains one CNN branch.
As can be seen in Table~\ref{table_QGAN}, QGAN performs a little worse than LUGAN.
We analyze, in addition to ensuring $Q$ being full-rank, LUGAN could also increase the nonlinearity of the model by multiplying $Q^L$ and $Q^U$.

\begin{table}[t]
	\small
	\centering
	\caption{Validation of the hypergraph convolution. Baseline-H$i$ ($i=1,2,3$) means that the gait features are learned by $i$-order HGC from the single view pose. And LUGAN-H$i$ learn features from complete-view pose sequences.}
	\begin{tabular}{c|ccc|c}
		\hline
		\multirow{2}{*}{Methods} & \multicolumn{3}{c|}{CASIA-B} & \multirow{2}{*}{OUMVLP-Pose} \\ \cline{2-4}
		& NM\#5-6  & BG\#1-2 & CL\#1-2 &                         \\ \hline
		Baseline-H1                 & 87.0     & 74.2    & 65.7    & 42.7                    \\
		Baseline-H2                 & 87.5     & 75.5     & 67.0    & 43.5                    \\
		Baseline-H3                 & 87.3     & 75.5    & 67.1    & 43.3                    \\ \hline
		LUGAN-H1                    & 87.1     & 78.4    & 73.8    & 46.8                    \\
		LUGAN-H2                    & 88.3     & 79.7    & 75.4    & 47.5                    \\
		LUGAN-H3                    & 87.5     & 79.0    & 75.3    & 47.9                    \\ \hline
	\end{tabular}
	\label{table_HGCN}
\end{table}

\subsubsection{Validation of HGC}
\label{VoH}
To verify the validity of high-level correlation learning, we conduct experiments under different sacles of hypergraphs.
Specifically, two groups of experimental settings are considered: 1) learning gait features from the single-view pose sequences, 2) learning gait features from the complete-view pose sequences.
The results are presented in Table~\ref{table_HGCN}.
As can be seen, the HGC-based architecture outperforms the GC-based (H1) one with or without multi-view pose generation.
Furthermore, we visualize the normalized adjacency matrix of hypergraph, i.e., $D^{-1/2} H B^{-1} H^T D^{-1/2}$, in Fig.~\ref{fig_adj}.
We found that the adjacency of higher-order hypergraph contains more non-zero elements.

\begin{figure*}[ht]
	\centering
	\subfigure[CASIA-B]{\includegraphics[width=0.45\textwidth]{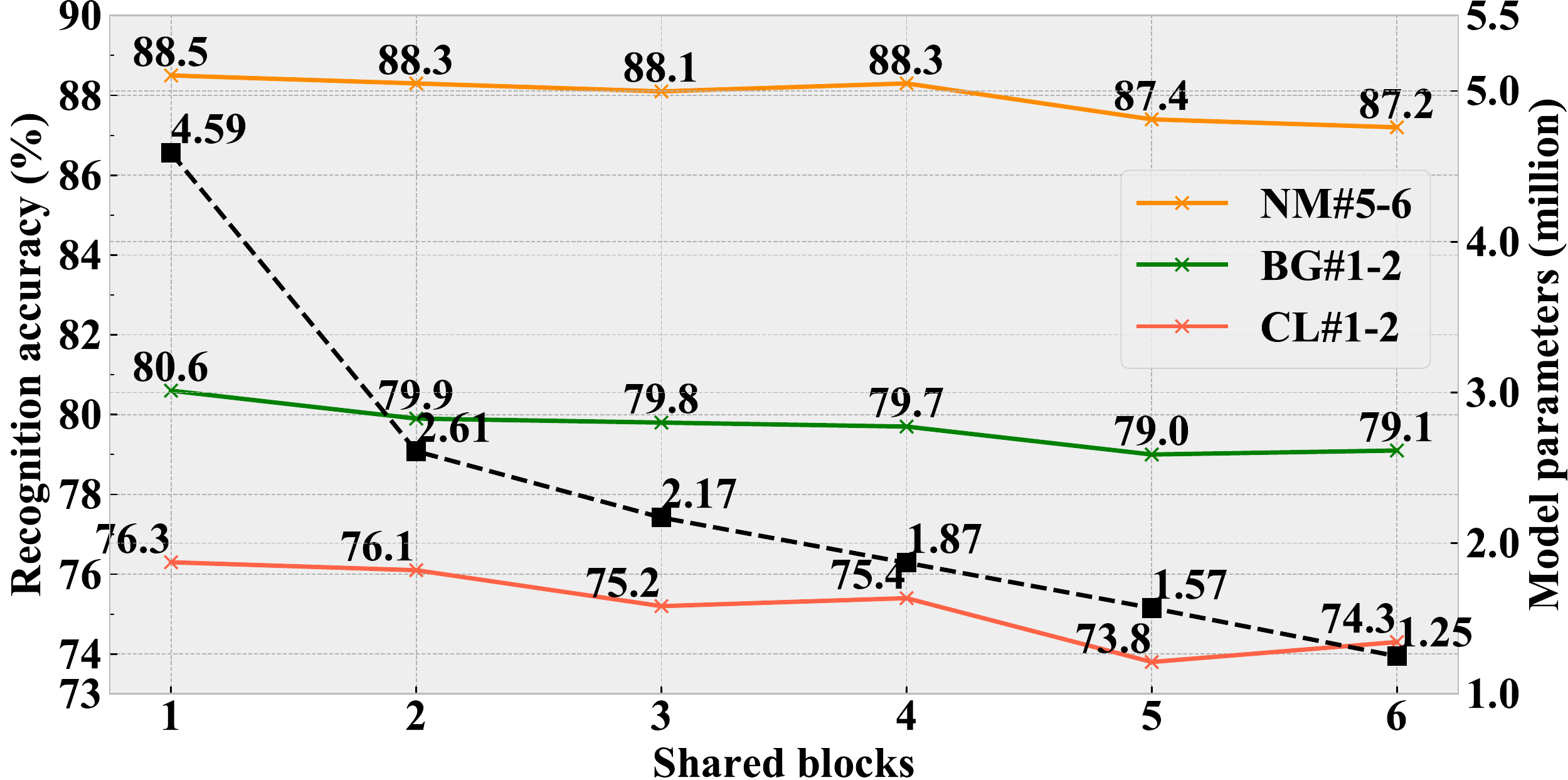}}
	\subfigure[OUMVLP-pose]{\includegraphics[width=0.45\textwidth]{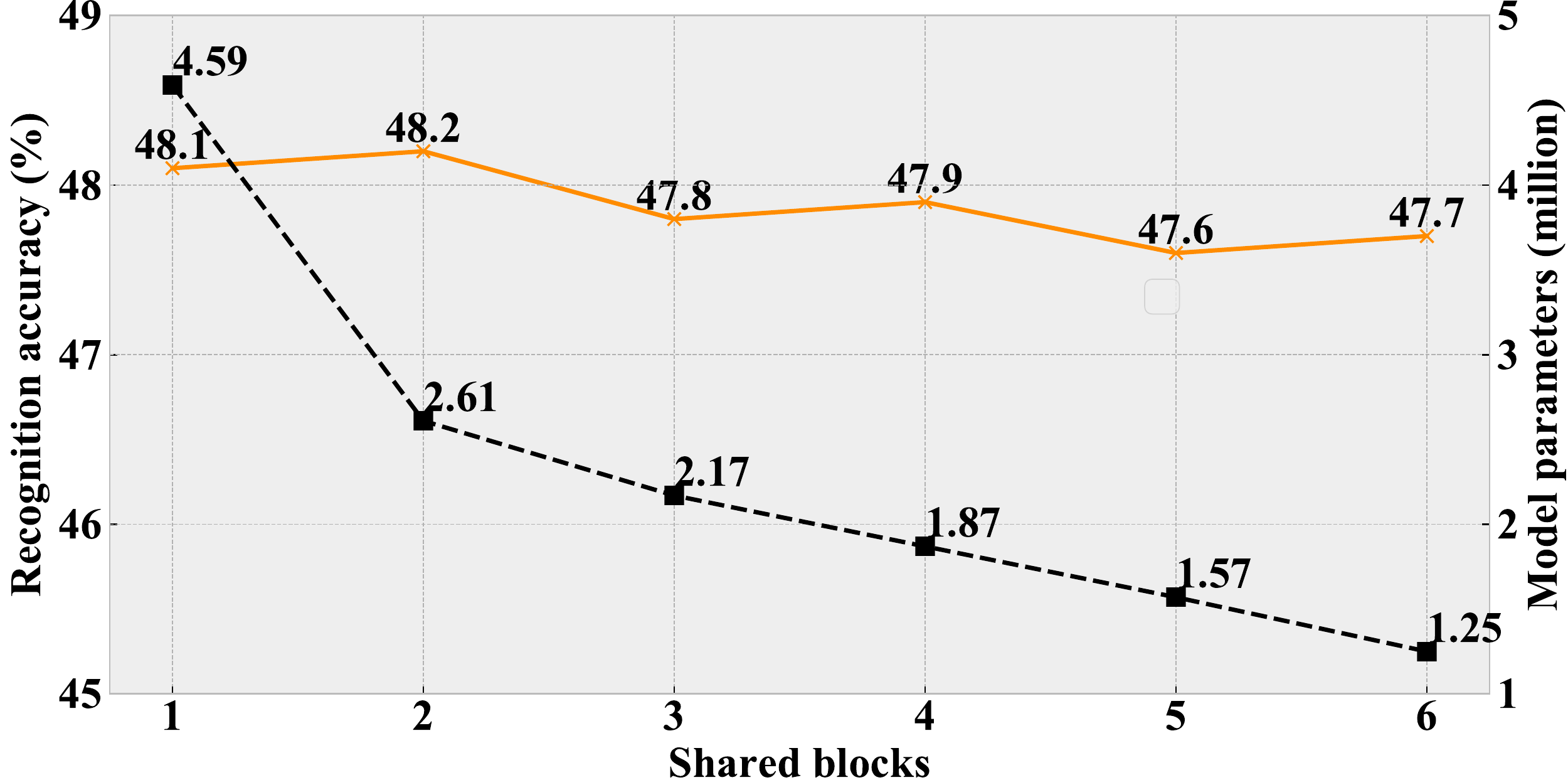}}
	\caption{Impact of the shared blocks, where the black squares denote the number of model parameters.}
	\label{fig_blocks}
\end{figure*}

\begin{table*}[ht]
	\small
	\centering
	\caption{Comparison with existing models on CASIA-B~\cite{CASIA-B}. The best performance of each subset is marked in \textcolor{red}{red}.}
	\begin{tabular}{cc|ccccccccccc|c}
		\hline
		\multicolumn{2}{c|}{Gallery NM\#1-4}                       & \multicolumn{11}{c|}{0°-180°}                                              & \multirow{2}{*}{mean} \\ \cline{1-13}
		\multicolumn{2}{c|}{Probe}                                 & 0°   & 18°  & 36°  & 54°  & 72°  & 90° & 108° & 126° & 144° & 162° & 180° &                       \\ \hline
		\multicolumn{1}{c|}{\multirow{8}{*}{NM\#5-6}} & PTSN~\cite{PTSN}       & 34.5 & 45.6 & 45.6 & 51.3 & 51.3 & 52.3 & 53.0 & 50.8 & 52.2 & 48.3 & 31.4 & 47.4                  \\
		\multicolumn{1}{c|}{}                         & PTSN-3D~\cite{PTSN}    & 38.7 & 50.2 & 55.9 & 56.0 & 56.7 & 54.6 & 54.8 & 56.0 & 54.1 & 52.4 & 40.2 & 51.9                  \\
		\multicolumn{1}{c|}{}                         & JointsGait~\cite{JointsGait} & 68.1 & 73.6 & 77.9 & 76.4 & 77.5 & 79.1 & 78.4 & 76.0 & 69.5 & 71.9 & 70.1 & 74.4                  \\
		\multicolumn{1}{c|}{}                         & PoseGait~\cite{PoseGait}   & 55.3 & 69.6 & 73.9 & 75.0 & 68.0 & 68.2 & 71.1 & 72.9 & 76.1 & 70.4 & 55.4 & 68.7                  \\
		\multicolumn{1}{c|}{}                         & GaitSet~\cite{GaitSet}  & 71.6 & 87.7 & \textcolor{red}{92.6} & 89.1 & 82.4 & 80.3 & 84.4 & \textcolor{red}{89.0} & \textcolor{red}{89.8} & 82.9 & 66.6 & 83.3                  \\
		\multicolumn{1}{c|}{}                         & GaitGraph~\cite{GaitGraph}  & 83.9 & \textcolor{red}{88.1} & 88.3 & 89.6 & 86.7 & 87.4	 & 86.7 & 87.4 & 88.2 & \textcolor{red}{88.4} & 82.3 & 87.0                  \\
		\multicolumn{1}{c|}{}                         & GaitGraph2~\cite{GaitGraph2}  & 78.5 & 82.9 & 85.8 & 85.6 & 83.1 & 81.5 & 84.3 & 83.2 & 84.2 & 81.6 & 71.8 & 82.0                  \\ \cline{2-14} 
		\multicolumn{1}{c|}{}                         & ours        & \textcolor{red}{89.3}     & \textcolor{red}{88.1}     & 89.0     & \textcolor{red}{89.9}     & \textcolor{red}{87.4}     & \textcolor{red}{88.7}     & \textcolor{red}{87.4}     & 88.8     & 88.8     & 87.0     & \textcolor{red}{87.0}     & \textcolor{red}{88.3}                    \\ \hline
		\multicolumn{1}{c|}{\multirow{8}{*}{BG\#1-2}} & PTSN~\cite{PTSN}       & 22.4 & 29.8 & 29.6 & 29.6 & 29.6 & 31.5 & 32.1 & 31.0 & 27.3 & 28.1 & 18.2 & 28.3                  \\
		\multicolumn{1}{c|}{}                         & PTSN-3D~\cite{PTSN}    & 27.7 & 32.7 & 37.4 & 35.0 & 37.1 & 37.5 & 37.7 & 36.9 & 33.8 & 31.8 & 27.0 & 34.1                  \\
		\multicolumn{1}{c|}{}                         & JointsGait~\cite{JointsGait} & 54.3 & 59.1 & 60.6 & 59.7 & 63.0 & 65.7 & 62.4 & 59.0 & 58.1 & 58.6 & 50.1 & 59.1                  \\
		\multicolumn{1}{c|}{}                         & PoseGait~\cite{PoseGait}   & 35.3 & 47.2 & 52.4 & 46.9 & 45.5 & 43.9 & 46.1 & 48.1 & 49.4 & 43.6 & 31.1 & 44.5                  \\
		\multicolumn{1}{c|}{}                         & GaitSet~\cite{GaitSet}  & 64.1 & 76.4 & 81.4 & \textcolor{red}{82.4} & 77.2 & 71.8 & 75.4 & 80.8 & 81.2 & 75.7 & 59.4 & 75.1                  \\
		\multicolumn{1}{c|}{}                         & GaitGraph~\cite{GaitGraph}  & 78.0 & 77.9 & 78.1 & 75.5 & 72.8 & 70.1 & 70.6 & 72.4 & 74.5 & 78.0 & 68.7 & 74.2                  \\
		\multicolumn{1}{c|}{}                         & GaitGraph2~\cite{GaitGraph2}  & 69.9 & 75.9 & 78.1 & 79.3 & 71.4 & 71.7 & 74.3 & 76.2 & 73.2 & 73.4 & 61.7 & 73.2                  \\ \cline{2-14} 
		\multicolumn{1}{c|}{}                         & ours       & \textcolor{red}{79.4}     & \textcolor{red}{79.5}     & \textcolor{red}{71.6}     & \textcolor{red}{82.4}     & \textcolor{red}{78.1}     & \textcolor{red}{76.2}     & \textcolor{red}{78.7}     & \textcolor{red}{82.0}     & \textcolor{red}{81.6}     & \textcolor{red}{83.0}     & \textcolor{red}{73.6}     & \textcolor{red}{79.7}                     \\ \hline
		\multicolumn{1}{c|}{\multirow{8}{*}{CL\#1-2}} & PTSN~\cite{PTSN}       & 14.2 & 17.1 & 17.6 & 19.3 & 19.5 & 20.0 & 20.1 & 17.3 & 16.5 & 18.1 & 14.0 & 17.6                  \\
		\multicolumn{1}{c|}{}                         & PTSN-3D~\cite{PTSN}    & 15.8 & 17.2 & 19.9 & 20.0 & 22.3 & 24.3 & 28.1 & 23.8 & 20.9 & 23.0 & 17.0 & 21.1                  \\
		\multicolumn{1}{c|}{}                         & JointsGait~\cite{JointsGait} & 48.1 & 46.9 & 19.6 & 50.5 & 51.0 & 52.3 & 49.0 & 46.0 & 48.7 & 53.6 & 52.0 & 49.8                  \\
		\multicolumn{1}{c|}{}                         & PoseGait~\cite{PoseGait}   & 24.3 & 29.7 & 41.3 & 38.8 & 38.2 & 38.5 & 41.6 & 44.9 & 42.2 & 33.4 & 22.5 & 36.0                  \\
		\multicolumn{1}{c|}{}                         & GaitSet~\cite{GaitSet}  & 36.4 & 49.7 & 54.6 & 49.7 & 48.7 & 45.2 & 45.5 & 48.2 & 47.2 & 41.4 & 30.6 & 45.2                  \\
		\multicolumn{1}{c|}{}                         & GaitGraph~\cite{GaitGraph}  & 62.0 & 60.8 & 64.3 & 65.8 & 67.8 & 65.7 & 70.7 & 63.8 & 67.5 & 68.7 & 65.0 & 65.7                  \\
		\multicolumn{1}{c|}{}                         & GaitGraph2~\cite{GaitGraph2}  & 57.1 & 61.1 & 68.9 & 66.0 & 67.8 & 65.4 & 68.1 & 67.2 & 63.7 & 63.6 & 50.4 & 63.6                  \\ \cline{2-14} 
		\multicolumn{1}{c|}{}                         & ours        & \textcolor{red}{72.8}     & \textcolor{red}{72.3}     & \textcolor{red}{69.4}     & \textcolor{red}{75.2}     & \textcolor{red}{77.0}     & \textcolor{red}{79.6}     & \textcolor{red}{80.5}     & \textcolor{red}{78.1}     & \textcolor{red}{76.3}     & \textcolor{red}{74.9}     & \textcolor{red}{72.8}     & \textcolor{red}{75.4}                      \\ \hline
	\end{tabular}
	\label{table_compare_casia}
\end{table*}

\subsubsection{Exploration of the Framework Structure}
In the framework of Fig.~\ref{fig_framework}, the generative branch contains 7 blocks: part of them are parameter-independent and the rest share weights.
In this section, we explore the impact of shared blocks (1$\sim$6) to the accuracy and model size.
As shown in Fig.~\ref{fig_blocks}, even though the more shared blocks contribute to less model parameters, it also leads to the worse performance, especially for 5 and 6 shared blocks.
And we found that it is a trade-off to set 4 shared blocks, which gains a promising accuracy with the least parameters.

\subsection{Comparison and Visualization}
\begin{figure}[!t]
	\centering
	\subfigure[baseline]{\includegraphics[width=0.24\textwidth]{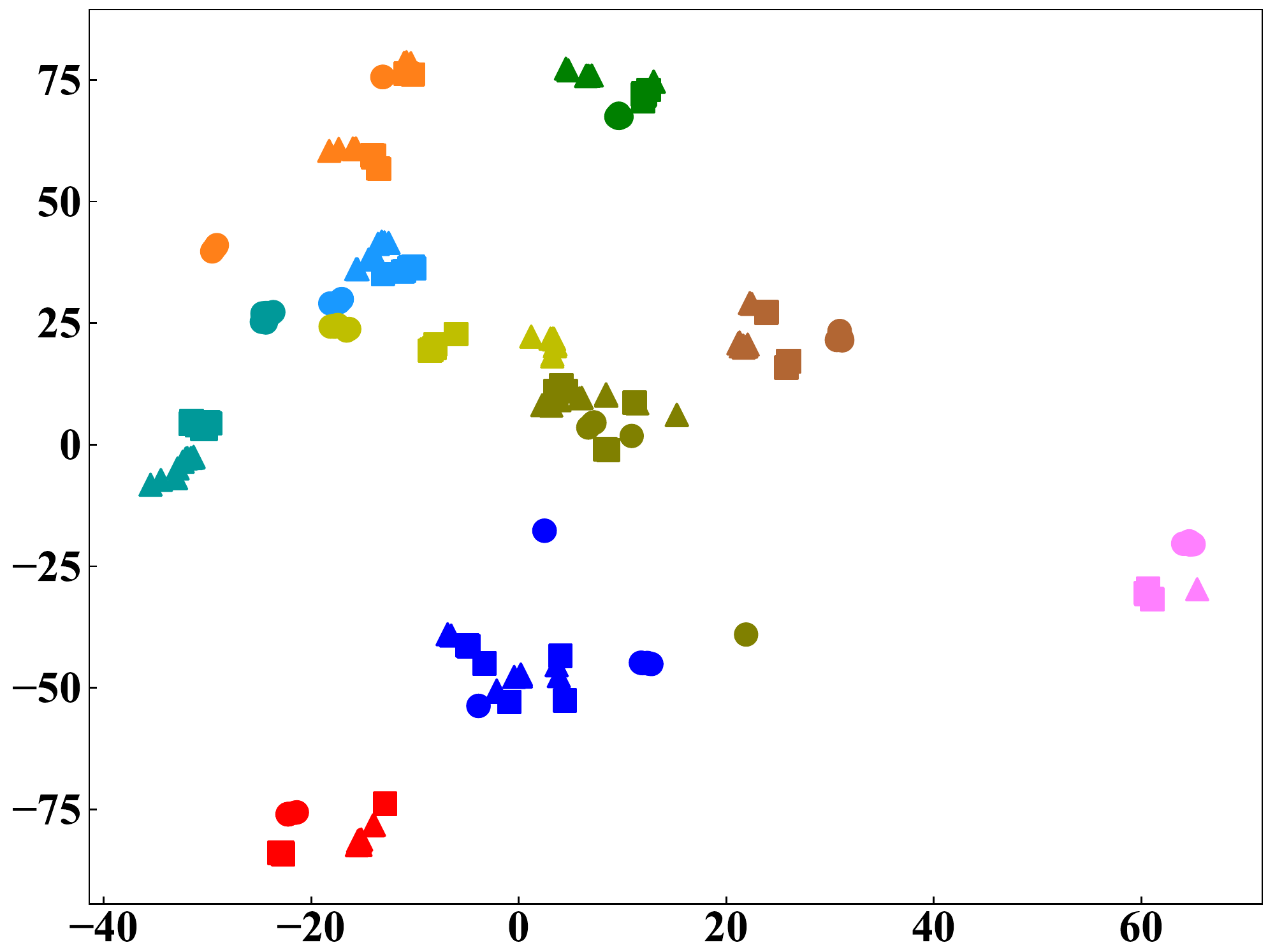}}
	\subfigure[LUGAN]{\includegraphics[width=0.24\textwidth]{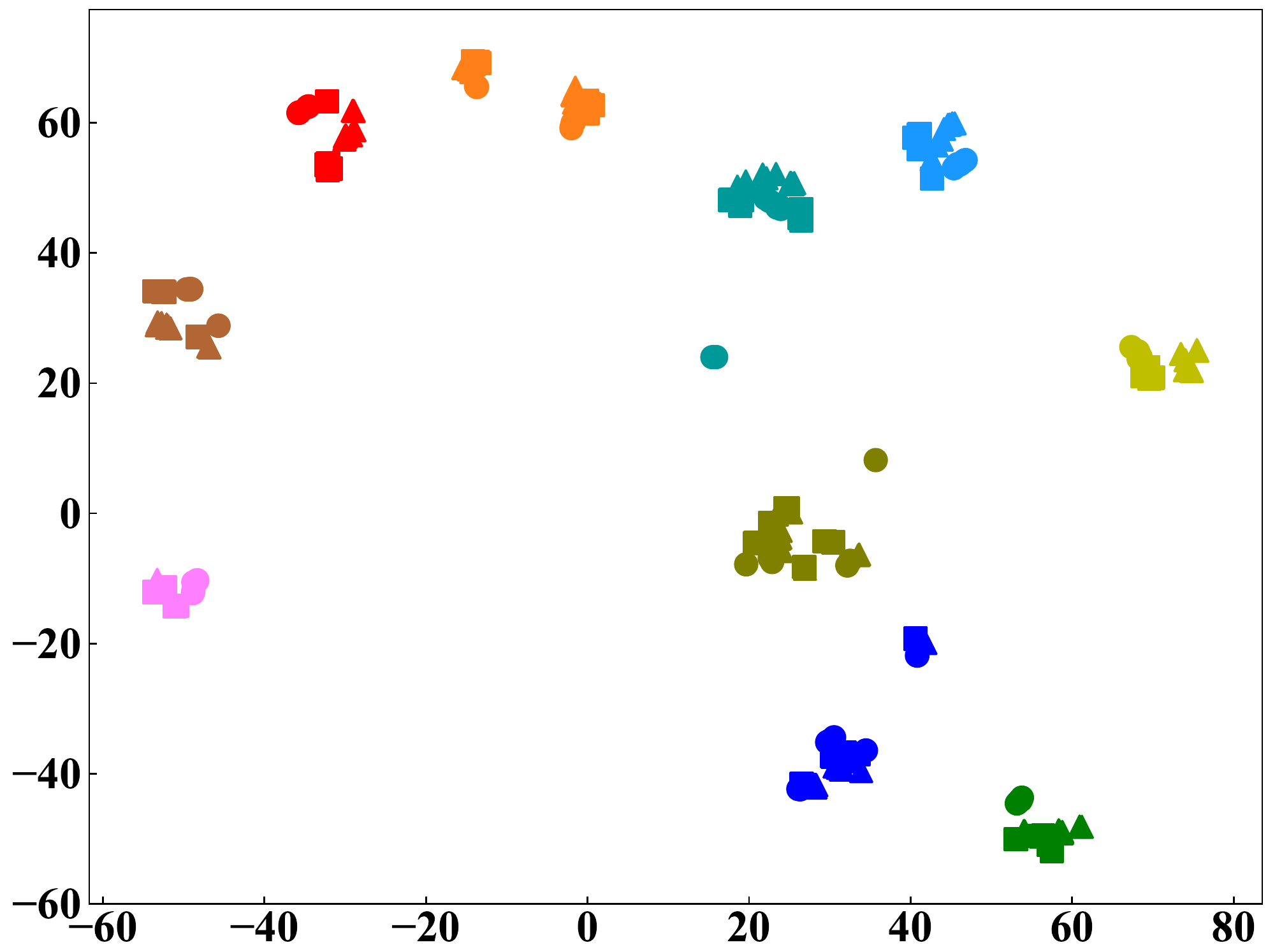}}
	\caption{Visualization of the learned embeddings via t-SNE~\cite{tsne}. Marks of the same color denote the same identity; marks of the same shape (circle, square and triangle) correspond to the same camera view.
	}
	\label{fig_tsne}
\end{figure}

In this section, we compare our method with existing model-based gait recognition methods on CASIA-B~\cite{CASIA-B} and OUMVLP-Pose~\cite{OU-ISIR}, and report the results in Table~\ref{table_compare_casia} and Table~\ref{table_compare_oumvlp}, respectively.
We conclude that our model gains the state-of-the-art performance.
Moreover, we visualize the learned gait embeddings in Fig.~\ref{fig_tsne}.
Specifically, we randomly choose 10 identities from CASIA-B~\cite{CASIA-B} and perform dimensional reduction on their features via t-SNE~\cite{tsne}.
We exhibit features of three camera views: 18° (circle), 90° (triangle) and 144° (square).
As can be seen, features with the multi-view generation have a larger inter-class variance and a smaller intra-class variance.

\begin{table*}[ht]
	\small
	\centering
	\caption{Comparison with existing models on OUMVLP-Pose~\cite{OU-ISIR}. The best performance is marked in \textcolor{red}{red}.}
	\begin{tabular}{c|ccccccc|ccccccc|c}
		\hline
		\multirow{2}{*}{Method} & \multicolumn{7}{c|}{0°-90°}                    & \multicolumn{7}{c|}{180°-270°}                 & \multirow{2}{*}{mean} \\ \cline{2-15}
		& 0°   & 15°  & 30°  & 45°  & 60°  & 75°  & 90° & 180° & 195° & 210° & 225° & 240° & 255° & 270° &                       \\ \hline
		CNN-Pose~\cite{OU-ISIR}                & 7.5 & 14.3 & 18.7 & 23.2 & 22.8 & 18.3 & 11.8 & 7.8 & 12.8 & 12.2 & 19.0 & 17.3 & 12.1 & 9.0 & 14.8                      \\
		GaitGraph~\cite{GaitGraph}                & 36.6 & 45.5 & 47.9 & 48.4 & 47.5 & 48.9 & 41.8 & 44.6 & 40.3 & 40.1 & 40.5 & 40.9 & 41.7 & 33.3 & 42.7                      \\
		GaitGraph2~\cite{GaitGraph2}                  & 32.9 & \textcolor{red}{47.7} & \textcolor{red}{53.9} & \textcolor{red}{56.8} & \textcolor{red}{53.9} & \textcolor{red}{54.7} & 45.4& 29.0 & 35.7 & 34.3 & 44.3 & 46.2 & 46.4 & 38.4 & 44.3                      \\ \hline
		ours               & \textcolor{red}{42.6} & \textcolor{red}{47.7} & 52.1 & 53.6 & 50.4 & 51.6 & \textcolor{red}{48.0} & \textcolor{red}{49.6} & \textcolor{red}{45.8} & \textcolor{red}{47.5} & \textcolor{red}{47.6} & \textcolor{red}{47.0} & \textcolor{red}{47.4} & \textcolor{red}{40.7} & \textcolor{red}{47.9}                      \\ \hline
	\end{tabular}
	\label{table_compare_oumvlp}
\end{table*}

\section{Conclusions}
\label{conclusion}
For the purpose of reducing cross-view pose variance and exploring high-level correlation, this paper proposed the LUGAN for multi-view pose generation and multi-scale hypergraph convolution.
For LUGAN, we exploited the spatial geometric relationship between cross-view pose and aimed to learn a full-rank matrix that transforms the source pose sequence to the target view.
To this end, the generator of LUGAN first encoded the source pose sequence and target view via GCN layers and FC layers, respectively, then learned a lower triangular matrix and an upper triangular matrix by two CNN branches, so that their matrix multiplication could be full rank.
And a condition discriminator was devised to train the generator.
For the high-level correlation learning, we proposed the 2-order and 3-order hypergraph convolution to respectively model the correlations of part-level and body-level, which were inserted as a differentiable plugin in the model.
Extensive experiments on two large pose-based gait datasets demonstrated the validity of these two improvements.

	\bibliographystyle{IEEEtran}
	\bibliography{IEEEabrv,mybibfile}

\begin{thebibliography}{10}
\providecommand{\url}[1]{#1}
\csname url@samestyle\endcsname
\providecommand{\newblock}{\relax}
\providecommand{\bibinfo}[2]{#2}
\providecommand{\BIBentrySTDinterwordspacing}{\spaceskip=0pt\relax}
\providecommand{\BIBentryALTinterwordstretchfactor}{4}
\providecommand{\BIBentryALTinterwordspacing}{\spaceskip=\fontdimen2\font plus
\BIBentryALTinterwordstretchfactor\fontdimen3\font minus
  \fontdimen4\font\relax}
\providecommand{\BIBforeignlanguage}[2]{{%
\expandafter\ifx\csname l@#1\endcsname\relax
\typeout{** WARNING: IEEEtran.bst: No hyphenation pattern has been}%
\typeout{** loaded for the language `#1'. Using the pattern for}%
\typeout{** the default language instead.}%
\else
\language=\csname l@#1\endcsname
\fi
#2}}
\providecommand{\BIBdecl}{\relax}
\BIBdecl

\bibitem{GaitGraph}
T.~Teepe, A.~Khan, and et~al., ``Gaitgraph: Graph convolutional network for
  skeleton-based gait recognition,'' in \emph{2021 IEEE International
  Conference on Image Processing}.\hskip 1em plus 0.5em minus 0.4em\relax IEEE,
  2021, pp. 2314--2318.

\bibitem{choi2019skeleton}
S.~Choi, J.~Kim, and et~al., ``Skeleton-based gait recognition via robust
  frame-level matching,'' \emph{IEEE Transactions on Information Forensics and
  Security}, vol.~14, no.~10, pp. 2577--2592, 2019.

\bibitem{zou2020deep}
Q.~Zou, Y.~Wang, and et~al., ``Deep learning-based gait recognition using
  smartphones in the wild,'' \emph{IEEE Transactions on Information Forensics
  and Security}, vol.~15, pp. 3197--3212, 2020.

\bibitem{GaitSurvey}
A.~Moghaddam and A.~Etemad, ``Deep gait recognition: A survey,'' \emph{arXiv
  preprint arXiv:2102.09546}, 2021.

\bibitem{TCDesc}
H.~Pan, Y.~Chen, and et~al., ``Tcdesc: Learning topology consistent descriptors
  for image matching,'' \emph{IEEE Transactions on Circuits and Systems for
  Video Technology}, vol.~32, no.~5, pp. 2845--2855, 2021.

\bibitem{MT3D}
B.~Lin, S.~Zhang, and F.~Bao, ``Gait recognition with multiple-temporal-scale
  3d convolutional neural network,'' in \emph{Proceedings of the 28th ACM
  International Conference on Multimedia}, 2020, pp. 3054--3062.

\bibitem{ACL}
Y.~Zhang, Y.~Huang, and et~al., ``Cross-view gait recognition by discriminative
  feature learning,'' \emph{IEEE Transactions on Image Processing}, vol.~29,
  pp. 1001--1015, 2019.

\bibitem{3DLocal}
Z.~Huang, D.~Xue, and et~al., ``3d local convolutional neural networks for gait
  recognition,'' in \emph{IEEE International Conference on Computer Vision},
  2021, pp. 14\,920--14\,929.

\bibitem{UGaitNet}
J.~Manuel, M.~Castro, and et~al., ``Ugaitnet: Multimodal gait recognition with
  missing input modalities,'' \emph{IEEE Transactions on Information Forensics
  and Security}, vol.~16, pp. 5452--5462, 2021.

\bibitem{tang2016robust}
J.~Tang, J.~Luo, and et~al., ``Robust arbitrary-view gait recognition based on
  3d partial similarity matching,'' \emph{IEEE Transactions on Image
  Processing}, vol.~26, no.~1, pp. 7--22, 2016.

\bibitem{PoseGait}
R.~Liao, S.~Y, and et~al., ``A model-based gait recognition method with body
  pose and human prior knowledge,'' \emph{Pattern Recognition}, vol.~98, p.
  107069, 2020.

\bibitem{JRCS}
P.~Limcharoen, N.~Khamsemanan, and C.~Nattee, ``View-independent gait
  recognition using joint replacement coordinates (jrcs) and convolutional
  neural network,'' \emph{IEEE Transactions on Information Forensics and
  Security}, vol.~15, pp. 3430--3442, 2020.

\bibitem{LGSD}
K.~Xu, X.~Jiang, and T.~Sun, ``Gait recognition based on local graphical
  skeleton descriptor with pairwise similarity network,'' \emph{IEEE
  Transactions on Multimedia}, 2021.

\bibitem{ResGCN}
Y.~Song, Z.~Zhang, and et~al., ``Stronger, faster and more explainable: A graph
  convolutional baseline for skeleton-based action recognition,'' in
  \emph{Proceedings of the 28th ACM International Conference on Multimedia},
  2020, pp. 1625--1633.

\bibitem{MGAN}
Y.~He, J.~Zhang, and et~al., ``Multi-task gans for view-specific feature
  learning in gait recognition,'' \emph{IEEE Transactions on Information
  Forensics and Security}, vol.~14, no.~1, pp. 102--113, 2018.

\bibitem{MvGGAN}
X.~Chen, X.~Luo, and et~al., ``Multi-view gait image generation for cross-view
  gait recognition,'' \emph{IEEE Transactions on Image Processing}, vol.~30,
  pp. 3041--3055, 2021.

\bibitem{CASIA-B}
S.~Yu, D.~Tan, and T.~Tan, ``A framework for evaluating the effect of view
  angle, clothing and carrying condition on gait recognition,'' in \emph{18th
  International Conference on Pattern Recognition}, vol.~4, 2006, pp. 441--444.

\bibitem{OU-ISIR}
W.~An, S.~Yu, and et~al., ``Performance evaluation of model-based gait on
  multi-view very large population database with pose sequences,'' \emph{IEEE
  Transactions on Biometrics, Behavior, and Identity Science}, vol.~2, no.~4,
  pp. 421--430, 2020.

\bibitem{GLConv}
B.~Lin, S.~Zhang, and X.~Yu, ``Gait recognition via effective global-local
  feature representation and local temporal aggregation,'' in \emph{IEEE
  Conference on Computer Vision and Pattern Recognition}, 2021, pp.
  14\,648--14\,656.

\bibitem{CSTL}
X.~Huang, D.~Zhu, and et~al., ``Context-sensitive temporal feature learning for
  gait recognition,'' in \emph{IEEE International Conference on Computer
  Vision}, 2021, pp. 12\,909--12\,918.

\bibitem{LSTM}
S.~Hochreiter and J.~Schmidhuber, ``Long short-term memory,'' \emph{Neural
  Computation}, vol.~9, no.~8, pp. 1735--1780, 1997.

\bibitem{GaitGAN}
S.~Yu, H.~Chen, and et~al., ``Gaitgan: Invariant gait feature extraction using
  generative adversarial networks,'' in \emph{IEEE Conference on Computer
  Vision and Pattern Recognition Workshops}, 2017, pp. 30--37.

\bibitem{CycleGAN}
J.~Zhu, T.~Park, and et~al., ``Unpaired image-to-image translation using
  cycle-consistent adversarial networks,'' in \emph{IEEE International
  Conference on Computer Vision}, 2017, pp. 2223--2232.

\bibitem{StarGAN}
Y.~Choi, M.~Choi, and et~al., ``Stargan: Unified generative adversarial
  networks for multi-domain image-to-image translation,'' in \emph{IEEE
  Conference on Computer Vision and Pattern Recognition}, 2018, pp. 8789--8797.

\bibitem{zhao20063d}
G.~Zhao, G.~Liu, and et~al., ``3d gait recognition using multiple cameras,'' in
  \emph{International Conference on Automatic Face and Gesture Recognition},
  2006, pp. 529--534.

\bibitem{JointsGait}
N.~Li, X.~Zhao, and C.~Ma, ``Jointsgait: A model-based gait recognition method
  based on gait graph convolutional networks and joints relationship pyramid
  mapping,'' \emph{arXiv preprint arXiv:2005.08625}, 2020.

\bibitem{GCN}
T.~Kipf and M.~Welling, ``Semi-supervised classification with graph
  convolutional networks,'' \emph{International Conference on Learning
  Representations}, pp. 1--14, 2017.

\bibitem{GCN1}
J.~Bruna, W.~Zaremba, and et~al., ``Spectral networks and locally connected
  networks on graphs,'' in \emph{International Conference on Learning
  Representations}, 2014, pp. 1--14.

\bibitem{GCN2}
M.~Defferrard, X.~Bresson, and P.~Gheynst, ``Convolutional neural networks on
  graphs with fast localized spectral filtering,'' in \emph{Advances in Neural
  Information Processing Systems}, 2016, pp. 3844--3852.

\bibitem{HGCN1}
S.~Bai, F.~Zhang, and H.~Torr, ``Hypergraph convolution and hypergraph
  attention,'' \emph{Pattern Recognition}, vol. 110, p. 107637, 2021.

\bibitem{HGCN2}
Y.~Feng, H.~You, and et~al., ``Hypergraph neural networks,'' in
  \emph{Proceedings of the AAAI Conference on Artificial Intelligence},
  vol.~33, no.~01, 2019, pp. 3558--3565.

\bibitem{ST-GCN}
J.~Yang, W.~Zheng, and et~al., ``Spatial-temporal graph convolutional network
  for video-based person re-identification,'' in \emph{IEEE Conference on
  Computer Vision and Pattern Recognition}, 2020, pp. 3289--3299.

\bibitem{MGH}
Y.~Yan, J.~Qin, and et~al., ``Learning multi-granular hypergraphs for
  video-based person re-identification,'' in \emph{IEEE Conference on Computer
  Vision and Pattern Recognition}, 2020, pp. 2899--2908.

\bibitem{AAGCN}
H.~Pan, Y.~Bai, and et~al., ``Aagcn: Adjacency-aware graph convolutional
  network for person re-identification,'' \emph{Knowledge-Based Systems}, p.
  107300, 2021.

\bibitem{STGCN}
S.~Yan, Y.~Xiong, and D.~Lin, ``Spatial temporal graph convolutional networks
  for skeleton-based action recognition,'' in \emph{Proceedings of the AAAI
  Conference on Artificial Intelligence}, 2018.

\bibitem{Shift-GCN}
K.~Cheng, Y.~Zhang, and et~al., ``Skeleton-based action recognition with shift
  graph convolutional network,'' in \emph{IEEE Conference on Computer Vision
  and Pattern Recognition}, 2020, pp. 183--192.

\bibitem{GAN}
I.~Goodfellow, J.~Pouget-Abadie, and et~al., ``Generative adversarial nets,''
  \emph{Advances in Neural Information Processing Systems}, vol.~27, 2014.

\bibitem{WGAN}
M.~Arjovsky and L.~Bottou, ``Towards principled methods for training generative
  adversarial networks,'' \emph{arXiv preprint arXiv:1701.04862}, 2017.

\bibitem{CGAN}
M.~Mirza and S.~Osindero, ``Conditional generative adversarial nets,''
  \emph{arXiv preprint arXiv:1411.1784}, 2014.

\bibitem{SGAN}
A.~Odena, ``Semi-supervised learning with generative adversarial networks,''
  \emph{arXiv preprint arXiv:1606.01583}, 2016.

\bibitem{PN-GAN}
X.~Qian, Y.~Fu, and et~al., ``Pose-normalized image generation for person
  re-identification,'' in \emph{Proceedings of the European Conference on
  Computer Vision}, 2018, pp. 650--667.

\bibitem{CMReID1}
P.~Dai, R.~Ji, and et~al., ``Cross-modality person re-identification with
  generative adversarial training.'' in \emph{International Joint Conference on
  Artificial Intelligence}, vol.~1, 2018, p.~2.

\bibitem{CMReID2}
G.~Wang, T.~Zhang, and et~al., ``Cross-modality paired-images generation for
  rgb-infrared person re-identification,'' in \emph{Proceedings of the AAAI
  Conference on Artificial Intelligence}, vol.~34, no.~07, 2020, pp.
  12\,144--12\,151.

\bibitem{HRNet}
K.~Sun, B.~Xiao, and et~al., ``Deep high-resolution representation learning for
  human pose estimation,'' in \emph{IEEE Conference on Computer Vision and
  Pattern Recognition}, 2019, pp. 5693--5703.

\bibitem{Openpose}
Z.~Cao, T.~Simon, and et~al., ``Realtime multi-person 2d pose estimation using
  part affinity fields,'' in \emph{IEEE Conference on Computer Vision and
  Pattern Recognition}, 2017, pp. 7291--7299.

\bibitem{SCL}
P.~Khosla, P.~Teterwak, and et~al., ``Supervised contrastive learning,''
  \emph{Advances in Neural Information Processing Systems}, vol.~33, 2020.

\bibitem{Adam}
D.~Kingma and J.~Ba, ``Adam: A method for stochastic optimization,''
  \emph{arXiv preprint arXiv:1412.6980}, 2014.

\bibitem{GAST-Net}
J.~Liu, J.~Rojas, and et~al., ``A graph attention spatio-temporal convolutional
  network for 3d human pose estimation in video,'' in \emph{2021 IEEE
  International Conference on Robotics and Automation (ICRA)}, 2021, pp.
  3374--3380.

\bibitem{PTSN}
R.~Liao, C.~Cao, and et~al., ``Pose-based temporal-spatial network (ptsn) for
  gait recognition with carrying and clothing variations,'' in \emph{Chinese
  Conference on Biometric Recognition}, 2017, pp. 474--483.

\bibitem{GaitSet}
H.~Chao, K.~Wang, and et~al., ``Gaitset: Cross-view gait recognition through
  utilizing gait as a deep set,'' \emph{IEEE Transactions on Pattern Analysis
  and Machine Intelligence}, 2021.

\bibitem{GaitGraph2}
T.~Teepe, J.~Gilg, and et~al., ``Towards a deeper understanding of
  skeleton-based gait recognition,'' in \emph{Proceedings of the IEEE/CVF
  Conference on Computer Vision and Pattern Recognition}, 2022, pp. 1569--1577.

\bibitem{tsne}
M.~Van and G.~Hinton, ``Visualizing data using t-sne.'' \emph{Journal of
  machine learning research}, vol.~9, no.~11, 2008.

\end{thebibliography}

\end{document}